%% file: AnonymousSubmission2027.tex
\documentclass[letterpaper]{article} 
\usepackage{aaai2027}  
\usepackage[hyphens]{url}  
\usepackage{graphicx} 
\usepackage{caption} 
\usepackage{algorithm}
\usepackage{algorithmic}

\usepackage[numbers]{natbib}

\usepackage{newfloat}
\DeclareCaptionStyle{ruled}{labelfont=normalfont,labelsep=colon,strut=off} 
 \usepackage{listings}
\floatstyle{ruled}
\newfloat{listing}{tb}{lst}{}
\floatname{listing}{Listing}
\floatname{listing}{Listing}
\usepackage{booktabs}
\usepackage{amssymb}
\usepackage{multirow} 
\usepackage{tabularx}
\usepackage{array}
\usepackage{booktabs}
\usepackage{multirow}

\usepackage{xcolor}
\usepackage{pifont}

\usepackage{booktabs}
\usepackage{multirow}
\usepackage{pgfplots}
\usepackage[most]{tcolorbox}
\usepackage{listings}

\newcommand{\rawhref}[2]{%
  \leavevmode
  \pdfstartlink
    attr{/Border [0 0 0]}
    user{/Subtype /Link
         /A << /S /URI /URI (\pdfescapestring{#1}) >>}%
  #2%
  \pdfendlink
}

\pgfplotsset{compat=1.18}

\newtcblisting{promptbox}[1]{
  title=#1,
  breakable,
  colback=gray!3,
  colframe=gray!45,
  boxrule=0.5pt,
  arc=2pt,
  left=6pt,
  right=6pt,
  top=6pt,
  bottom=6pt,
  fonttitle=\bfseries,
  listing only,
  listing options={
    basicstyle=\ttfamily\footnotesize,
    breaklines=true,
    breakatwhitespace=true,
    breakindent=0pt,
    columns=flexible,
    keepspaces=true,
    showstringspaces=false
  }
}

\newcommand{\promptinput}[4]{%
\begin{tcolorbox}[
  title={#1},
  breakable,
  colback=gray!3,
  colframe=gray!45,
  boxrule=0.5pt,
  arc=2pt,
  left=6pt,
  right=6pt,
  top=6pt,
  bottom=6pt,
  fonttitle=\bfseries
]
\lstinputlisting[
  firstline=#3,
  lastline=#4,
  basicstyle=\ttfamily\footnotesize,
  breaklines=true,
  breakatwhitespace=true,
  breakindent=0pt,
  columns=flexible,
  keepspaces=true,
  showstringspaces=false,
  literate={✅}{{[OK]}}4
           {❌}{{[BAD]}}5
           {→}{{$\rightarrow$}}1
           {≤}{{$\leq$}}1
           {·}{{$\cdot$}}1
           {—}{{--}}1
           {–}{{--}}1
           {“}{{``}}1
           {”}{{''}}1
]{#2}
\end{tcolorbox}
}

\newcommand{\cmark}{\textcolor{green!60!black}{\ding{51}}}
\newcommand{\xmark}{\textcolor{red!75!black}{\ding{55}}}
\nocopyright

\title{MultiRef-Compass: Towards Comprehensive Evaluation of Multi-Reference-to-Audio-Video Generation}
\author {
    Xiaohan Zhang\textsuperscript{\rm 1,\rm 2},
    Yuqing Wen\textsuperscript{\rm 3},
    Junlin Chen\textsuperscript{\rm 1}, 
    Yuqi Tang\textsuperscript{\rm 4}, 
    Yiting He\textsuperscript{\rm 5}, 
    Lizhuo Shao\textsuperscript{\rm 1}, 
    \\
    Weiming Zhu\textsuperscript{\rm 1},
    Tengfei Liu\textsuperscript{\rm 6}, 
    Yang Shi\textsuperscript{\rm 6,\rm 2}\corresponding,
    Jialu Chen\textsuperscript{\rm 2},
    Yuanxing Zhang\textsuperscript{\rm 2},
    Huaxiong Li\textsuperscript{\rm 1}\corresponding
}
\affiliations{
    $^1$Nanjing University\quad
    $^2$Kling Team\quad
    $^3$National University of Singapore\quad
    \\
    $^4$Hong Kong University of Science and Technology (Guangzhou)\quad
    $^5$Tsinghua University\quad
    $^6$Peking University\quad
    \\[3pt]
    \rawhref{https://github.com/zxhhh0201/MultiRef-Compass}{\textbf{Code}}
    \quad | \quad
    \rawhref{https://huggingface.co/datasets/zxhhhhhh/MultiRef-Compass}{\textbf{Dataset}}
    }

\begin{document}

\maketitle

\begin{abstract}
Multi-reference-to-audio-video (MR2AV) generation aims to generate coherent audio-video content conditioned on multiple references and textual instructions. 
Existing benchmarks mainly focus on text-driven generation, single-reference subject preservation, or isolated audio-video alignment, leaving the emerging MR2AV setting largely unexplored.
Compared with these settings, MR2AV requires models to jointly reason over multiple references while generating synchronized visual and audio content. 
Models must not only preserve each reference faithfully but also correctly bind and compose multiple referenced entities into coherent audio-visual events.
To address this gap, we introduce \textbf{MultiRef-Compass}, a unified benchmark for MR2AV generation. 
It comprises $350$ carefully curated samples constructed through a scalable and controllable asset-composition pipeline, covering multi-view subject preservation, multi-entity binding, and human-object-scene composition. 
To provide interpretable assessment, MultiRef-Compass defines an evaluation protocol with four dimensions: \textit{Basic Quality}, \textit{Reference Consistency}, \textit{Audio-Visual Consistency}, and \textit{Instruction Following}, using 14 sub-metrics. 
MultiRef-Compass integrates automatic metrics with a rejudging-enhanced MLLM-as-a-Judge framework, enabling scalable and auditable evaluation of both perceptual fidelity and reference-conditioned composition.
Extensive experiments on eight representative MR2AV systems reveal substantial room for improvement across multiple evaluation dimensions, underscoring the need for a comprehensive benchmark and positioning MultiRef-Compass as a foundation for future MR2AV research.
\end{abstract}

\begin{table*}[t]
\centering
\small

\begin{tabular}{@{}l c cc ccc cc@{}}
\toprule

\textbf{Benchmark}
& \textbf{\#Samples}
& \multicolumn{2}{c}{\textbf{Task}}
& \multicolumn{3}{c}{\textbf{Reference Configuration}}
& \multicolumn{2}{c}{\textbf{Reference Evaluation}}
\\

\cmidrule(lr){3-4}
\cmidrule(lr){5-7}
\cmidrule(lr){8-9}

&
&
\multirow{-2}{*}{\textbf{R2V}}
&
\multirow{-2}{*}{\textbf{X2AV}}
&
\multirow{-2}{*}{\textbf{Multi-Ref.}}
&
\shortstack{\textbf{Same-ID}\\\textbf{Multi-view}}
&
\shortstack{\textbf{Multi-modal}\\\textbf{Reference}}
&
\shortstack{\textbf{Reference}\\\textbf{Binding}}
&
\shortstack{\textbf{Multi-modal }\\\textbf{Consistency}}
\\

\midrule

T2AV-Compass~\cite{cao2025t2avcompass}
& 500
& \xmark
& \cmark
& \xmark
& \xmark
& \xmark
& \xmark
& \xmark
\\

AVGen-Bench~\cite{zhou2026avgenbench}
& 235
& \xmark
& \cmark
& \xmark
& \xmark
& \xmark
& \xmark
& \xmark
\\

AVBench~\cite{yang2026avbench}
& 470
& \xmark
& \cmark
& \xmark
& \xmark
& \xmark
& \xmark
& \xmark
\\

UniVBench~\citep{wei2026univbench}
& 200
& \cmark
& \xmark
& \cmark
& \xmark
& \xmark
& \xmark
& \xmark
\\

OpenS2V-Nexus~\cite{yuan2026opens2vnexus}
& 180
& \cmark
& \xmark
& \cmark
& \xmark
& \xmark
& \xmark
& \xmark
\\

\midrule

\textbf{MultiRef-Compass}
& \textbf{350}
& \textbf{\cmark}
& \textbf{\cmark}
& \textbf{\cmark}
& \textbf{\cmark}
& \textbf{\cmark\ (I/V/A)}
& \textbf{\cmark}
& \textbf{\cmark}
\\

\bottomrule
\end{tabular}
\caption{\textbf{Comparison between MultiRef-Compass and representative (audio-)video generation benchmarks.}
MultiRef-Compass focuses on MR2AV generation, where each reference can be represented by multiple images and combined with optional video and audio references. It requires models to effectively fuse diverse reference modalities, establish correct reference binding, and preserve multi-modal reference consistency.}
\label{tab:benchmark_comparison}
\end{table*}


\section{Introduction}
Recent video generation systems~\cite{seedance2026seedance,team2025kling,li2026skyreelsv3} increasingly support reference-based control beyond a single text prompt or a single image.
In practical content creation workflows, users often provide multiple images to specify the same subject from different viewpoints, define object and scene references, and describe interactions among multiple entities.
This trend has given rise to multi-reference-to-audio-video (MR2AV) generation, in which models must jointly interpret multiple references and textual instructions to produce coherent audio-video content.

Despite this growing capability, existing benchmarks remain insufficient for evaluating MR2AV systems.
Text-to-audio-video (T2AV) benchmarks~\cite{cao2025t2avcompass,wei2026univbench,liu2026longavcompass} primarily assess instruction following and audio-visual alignment, but generally do not require outputs to be grounded in multiple references. 
Existing reference-to-audio-video (R2AV) benchmarks~\cite{yuan2026opens2vnexus} emphasize subject consistency and reference fidelity, yet are largely designed for single-reference settings. 
Consequently, current evaluation protocols provide limited insight into whether models can correctly interpret multiple references, bind them to the intended entities and events, and compose them into coherent audio-video outputs.

As summarized in Table~\ref{tab:benchmark_comparison}, existing benchmarks cover only X-to-audio-video (X2AV) generation, where X denotes any input modality, or limited R2V settings. 
Unified evaluation of MR2AV generation remains underexplored.
Such evaluation must examine how models understand, bind, and integrate multiple references while maintaining multimodal consistency.
First, models require \emph{cross-reference understanding}: they must infer whether multiple reference images correspond to different views of the same subject or to distinct entities that should appear together. 
A common failure is identity splitting, where multiple views of the same subject are generated as different identities. 
Second, models require \emph{compositional binding}: they must correctly associate referenced entities with prompt-specified roles, actions, and interactions across both visual and audio modalities. Failures include identity swaps, attribute leakage, and incorrect event or sound-source assignments.
Finally, models require \emph{natural visual integration}. 
High reference similarity alone is insufficient, as models may rigidly copy reference appearances and produce copy-and-paste artifacts that fail to blend naturally into the generated scene. 
Together, these challenges demonstrate that MR2AV evaluation should assess not only reference fidelity, but also cross-reference understanding, compositional binding, and natural visual integration.

\begin{figure*}[t]
\centering
\includegraphics[width=\textwidth]{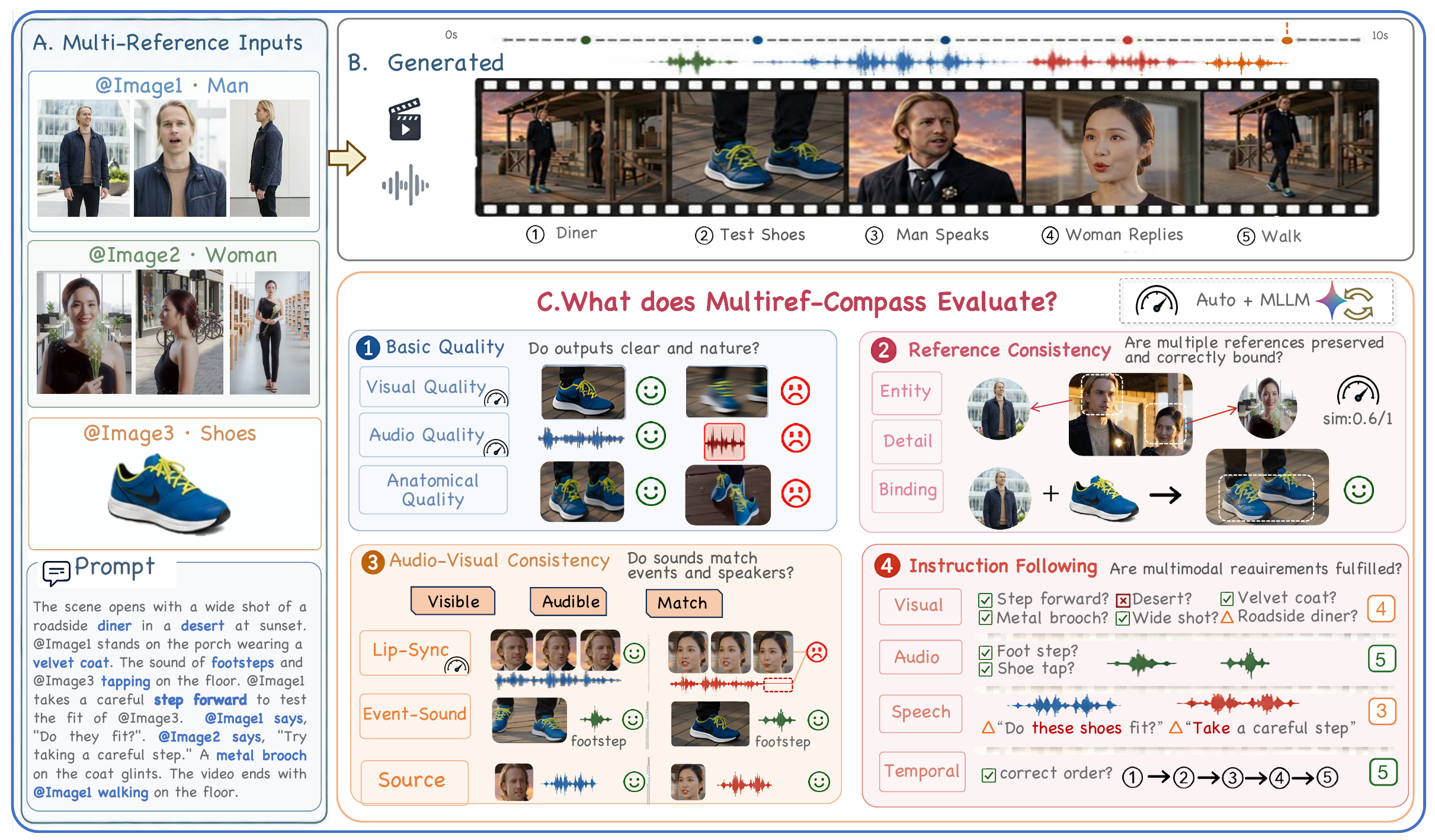}
\caption{\textbf{Overview of the MultiRef-Compass framework.} The benchmark evaluates MR2AV tasks through four complementary dimensions—Basic Quality, Reference Consistency, Audio-Visual Consistency, and Instruction Following, enabling fine-grained diagnosis of multi-reference-conditioned audio-visual generation failures beyond aggregate leaderboard scores.}
\label{fig:overall}
\end{figure*}

To address these challenges, as shown in Figure~\ref{fig:overall}, we introduce \textbf{MultiRef-Compass}, a comprehensive benchmark for MR2AV generation.  
MultiRef-Compass contains $350$ curated samples constructed through a scalable and controllable taxonomy-driven asset-composition pipeline, covering multi-view subject preservation, multi-entity binding, and human-object-scene composition.
The pipeline enables controlled composition of multimodal assets while maintaining reproducible sample construction. 
We further develop a hybrid evaluation framework that integrates automatic metrics with rejudging-enhanced MLLM-as-a-Judge evaluation, providing scalable and auditable assessment across four dimensions: \textit{Basic Quality}, \textit{Reference Consistency}, \textit{Audio-Visual Consistency}, and \textit{Instruction Following}. 
Moreover, MultiRef-Compass adopts an omni-reference schema with metric routing, making it readily extensible to image, video, and audio references alongside textual instructions.
Extensive experiments on eight representative MR2AV systems reveal substantial limitations of current models across multiple evaluation dimensions, demonstrating the effectiveness of MultiRef-Compass as a comprehensive benchmark for future MR2AV research.

Our contributions are summarized as follows:
\begin{itemize}
\item We introduce \textbf{MultiRef-Compass}, a comprehensive benchmark for MR2AV generation. It contains $350$ taxonomy-driven samples constructed through a scalable and controllable asset-composition pipeline, covering multi-view subject preservation, multi-entity binding, and human-object-scene composition.
\item We develop a hybrid evaluation framework that integrates automatic metrics with a rejudging-enhanced MLLM-as-a-Judge framework to evaluate MR2AV systems across four dimensions: \textit{Basic Quality}, \textit{Reference Consistency}, \textit{Audio-Visual Consistency}, and \textit{Instruction Following}.
\item We conduct extensive experiments on eight representative MR2AV systems, providing a comprehensive analysis of current model capabilities and revealing key limitations across multiple aspects of MR2AV generation.
\end{itemize}

\begin{figure*}[t]
\centering
\includegraphics[width=\textwidth]{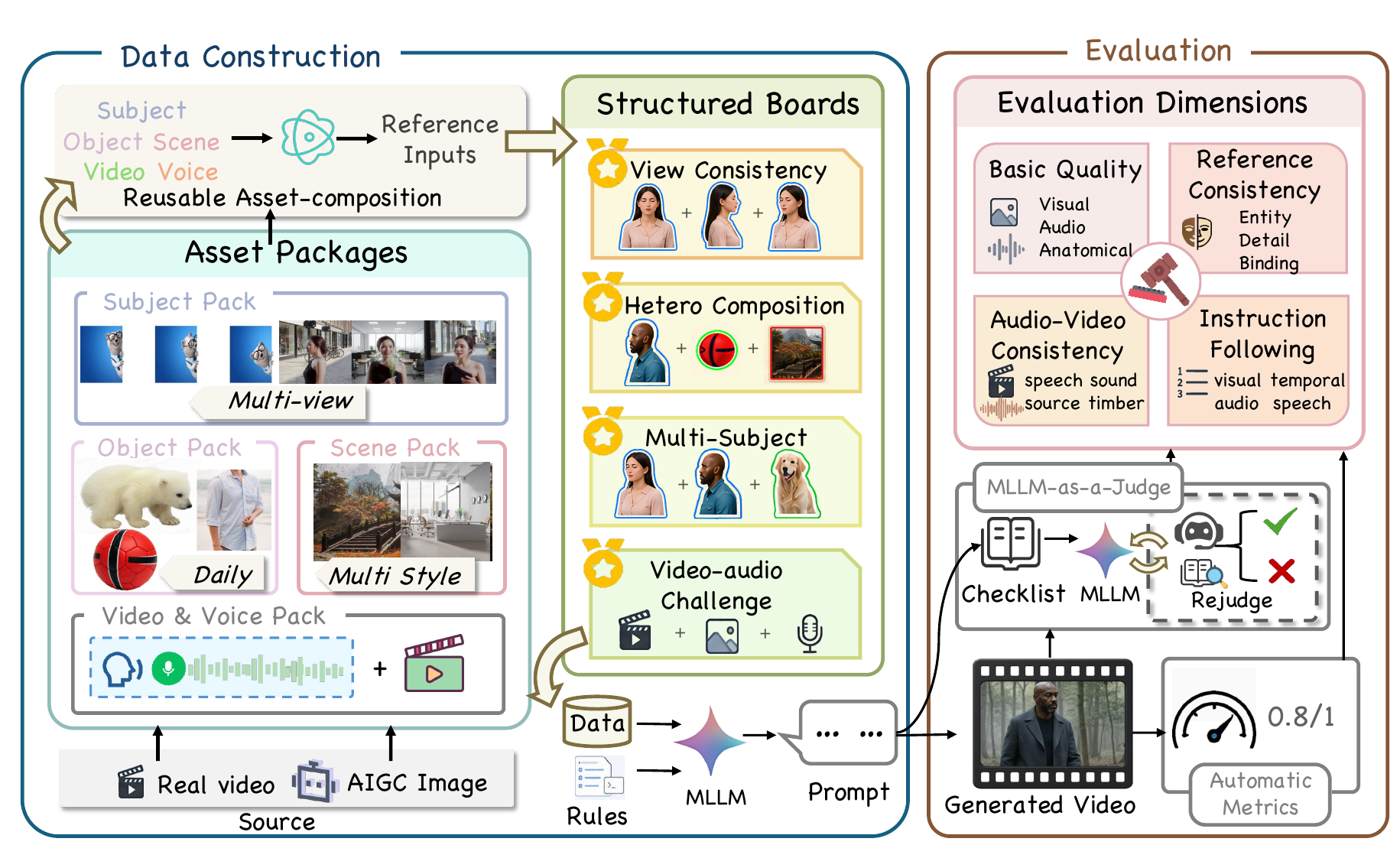}
\caption{\textbf{Data construction pipeline of MultiRef-Compass.} The benchmark is built from licensed asset packs covering subjects, objects, scenes, videos, and voices. These assets are filtered, annotated, and routed into different reference-conditioned tracks. After structured prompt generation and manual review, the evaluation framework combines automatic metrics with a rejudging-enhanced MLLM-as-a-Judge protocol, producing diagnostic scores across four major dimensions.}
\label{fig:pipeline}
\end{figure*}

\section{Related Work}

\subsection{Image-Conditioned Video Generation}

Image-to-video (I2V) generation has evolved from animating a single input image, often serving as an initial frame or key frame, to supporting more flexible reference-based control. Early methods focus on foundational capabilities of reference-based animation, including subject customization~\cite{jiang2023videobooth}, pose control~\cite{hu2023animateanyone}, identity consistency~\cite{xu2023magicanimate}, and localized motion~\cite{ma2024followyourclick}. Beyond these core capabilities, more advanced image-conditioned video generation methods incorporate richer spatial and motion cues, such as compositional spatiotemporal conditioning~\cite{wang2023videocomposer} and trajectory guidance~\cite{yin2023dragnuwa}, enabling more complex and fine-grained control over generated content.
More recent systems extend reference conditioning beyond a single image, allowing multiple images and heterogeneous modalities to specify the target generation elements~\cite{jiang2025vace,seedance2026seedance,team2025kling,li2026skyreelsv3}.
Unlike conventional I2V generation, these settings require models to determine relationships among multiple references and compose them according to textual instructions. 
This development motivates evaluation protocols that go beyond single-reference fidelity and assess cross-reference understanding, compositional binding, and multimodal consistency.

\subsection{Reference-Conditioned Generation Benchmarks}
Existing general-purpose video benchmarks, such as VBench~\cite{huang2023vbench}, VBench++~\cite{huang2024vbenchplusplus}, EvalCrafter~\cite{liu2023evalcrafter}, and FETV~\cite{liu2023fetv}, mainly assess generation quality and prompt alignment. 
UniVBench~\cite{wei2026univbench} covers multiple generation tasks, while OpenS2V-Nexus~\cite{yuan2026opens2vnexus} and UI2V-Bench~\cite{zhang2025ui2vbench} evaluate subject or reference consistency under image-conditioned settings. 
However, they rarely distinguish same-identity multi-view references from distinct-entity compositions.
Audio-video benchmarks introduce additional evaluation of audio quality, audio-visual alignment, event-sound correspondence, and synchronization. T2AV-Compass~\cite{cao2025t2avcompass} evaluates text-to-audio-video generation using automatic metrics and MLLM-based judgments; AVGen-Bench and AVBench further study multi-granular and human-aligned audio-video evaluation~\cite{zhou2026avgenbench,yang2026avbench}; and LongAV-Compass extends evaluation across long-form T2AV, I2AV, and V2AV settings~\cite{liu2026longavcompass}. 
Nevertheless, these benchmarks primarily consider text prompts or individual image/video conditions, providing limited diagnosis of how multiple references are jointly understood and bound to diverse multimodal generation conditions. 
MultiRef-Compass bridges these two lines of evaluation by jointly assessing multi-reference consistency and cross-modal alignment in audio-visual generation.

\section{MultiRef-Compass Dataset}
Multi-reference-to-audio-video (MR2AV) generation synthesizes new audio-video content from multiple references and a textual instruction. The references are not limited to images, but may include multimodal inputs, e.g., videos and audio clips, that provide information to be preserved or reflected in the generated content. The model must jointly interpret these references and compose them according to the instruction.
Figure~\ref{fig:pipeline} illustrates the overall dataset construction pipeline. 
\textbf{MultiRef-Compass} is built through three stages: (1) \textbf{asset-pack curation}, (2) \textbf{board-specific asset composition}, and (3) \textbf{structured prompt generation}. 
This design enables systematic control over reference types, entity combinations, audio conditions, and task difficulty, while ensuring reproducible benchmark construction.
\subsection{Data Construction}
\paragraph{Asset-Pack Curation.}
We curate scene, video, and voice packs primarily from royalty-free repositories (e.g., Pexels and Freesound), covering both real-world and cartoon styles. 
Subject packs are constructed through two pipelines: extracting multi-view identities from videos or generating $3$–$6$ auxiliary references from a primary image using AIGC tools, with variations in viewpoint, expression and pose. 
All references are manually verified for identity consistency, visibility, and visual quality. 
Object and scene packs span common foreground entities and diverse environments. 
Video and voice packs are further collected as optional subject and timbre references for the extended challenge evaluation.

\paragraph{Board-Specific Asset Composition.}
We then construct benchmark samples by recombining assets according to board-specific reference configurations. 
The main benchmark includes three boards. 
B1 draws multiple views from the same subject pack, representing same-identity multi-view conditioning.
B2 combines subject, object, and scene assets in different configurations to form heterogeneous compositions. 
B3 draws entities from multiple subject packs, creating multi-entity scenarios that require more complex interactions and role assignments. 
As an extended setting, B4 evaluates joint audio-video reference conditioning using both reference video and audio inputs, with results reported in the Appendix~\ref{app:challenge-board}.
Together, these boards provide systematic control over compositional complexity and cover a broad range of reference configurations. 
They evaluate whether models can correctly interpret reference relationships, bind entities, and integrate them into coherent audio-visual content.

\paragraph{Structured Prompt Generation.}
For each asset composition, we generate prompts using a structured schema rather than unconstrained free-form text. 
Conditioned on the predefined difficulty level, GPT-5.5~\cite{openai2026gpt55} first produces and refines three modality-specific fields describing the video, audio, and speech, and then integrates them into a unified generation prompt. 
We also create sample-level metadata that records evaluation-relevant conditions and enables the evaluation router to select the applicable metrics. 
Finally, human reviewers verify the grounding of the prompt in the selected assets, the plausibility of the described events, and the consistency among the prompt, metadata, and asset composition.

\subsection{Dataset Statistics}



\begin{figure*}[htbp]
\centering
\begin{minipage}[t]{0.25\textwidth}
\centering
\includegraphics[width=\linewidth]{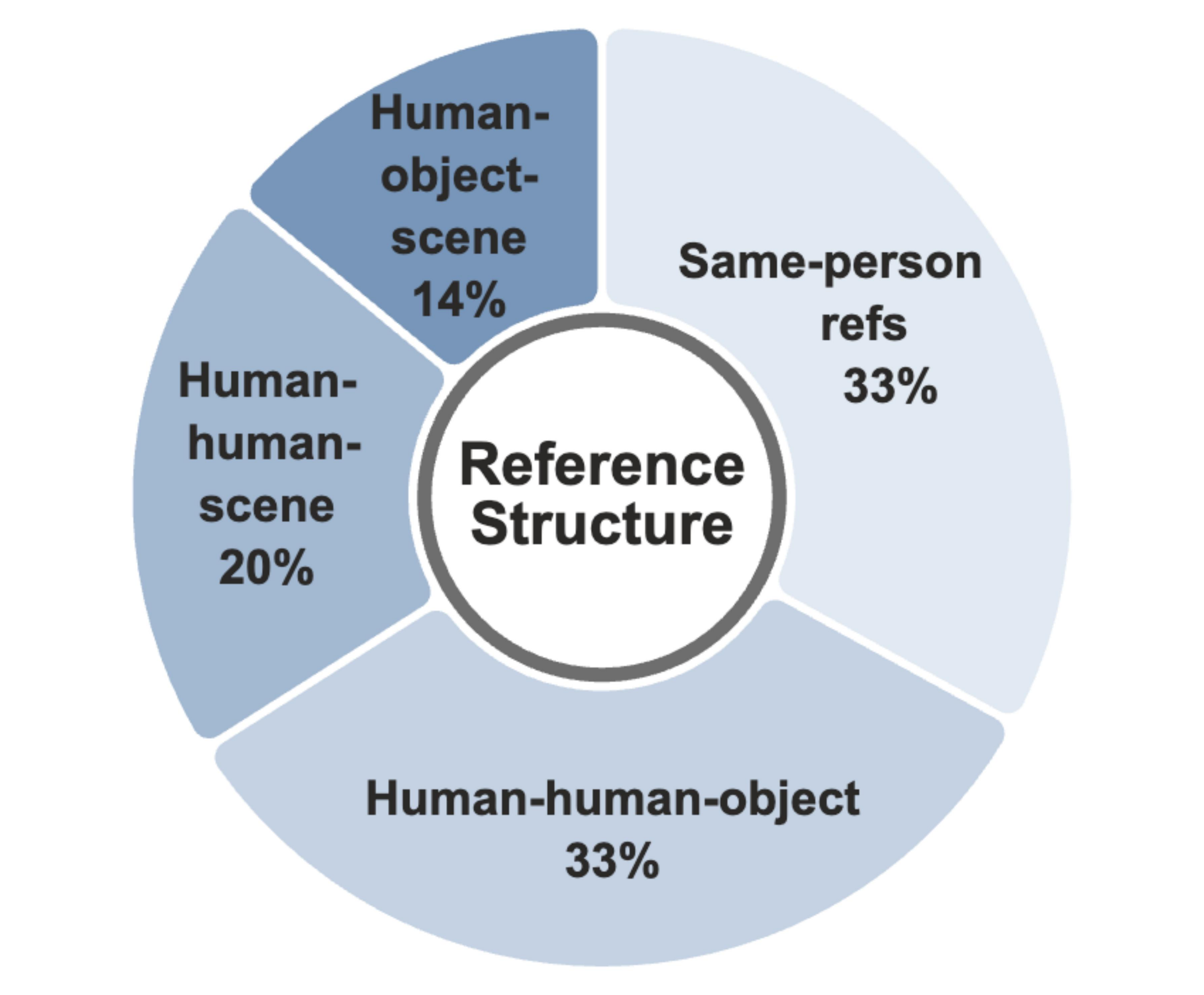}\\[-1mm]
{\footnotesize \textbf{(a)} Reference structure distribution}
\end{minipage}\hfill
\begin{minipage}[t]{0.25\textwidth}
\centering
\includegraphics[width=\linewidth]{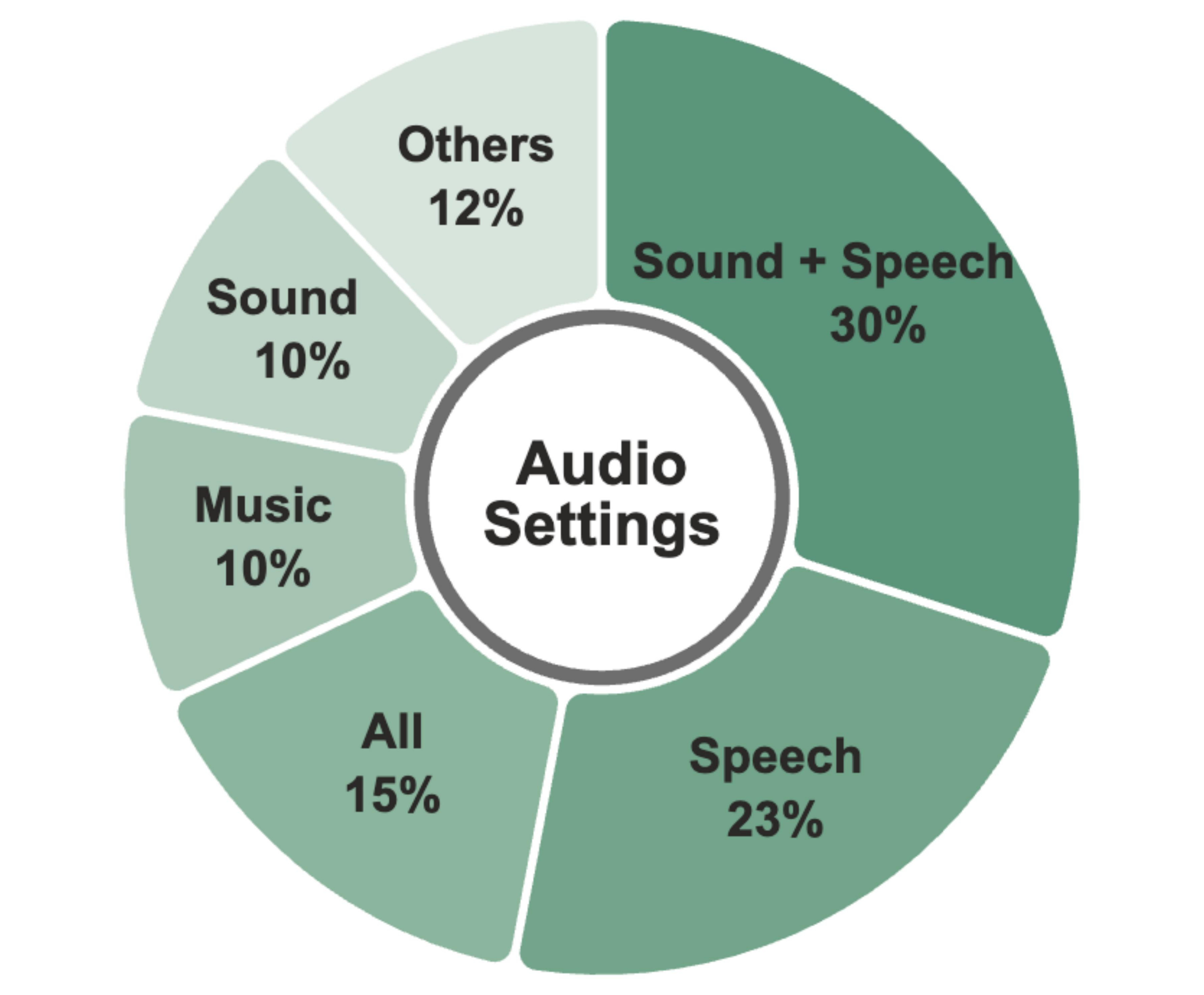}\\[-1mm]
{\footnotesize \textbf{(b)} Audio type distribution}
\end{minipage}\hfill
\begin{minipage}[t]{0.25\textwidth}
\centering
\includegraphics[width=\linewidth]{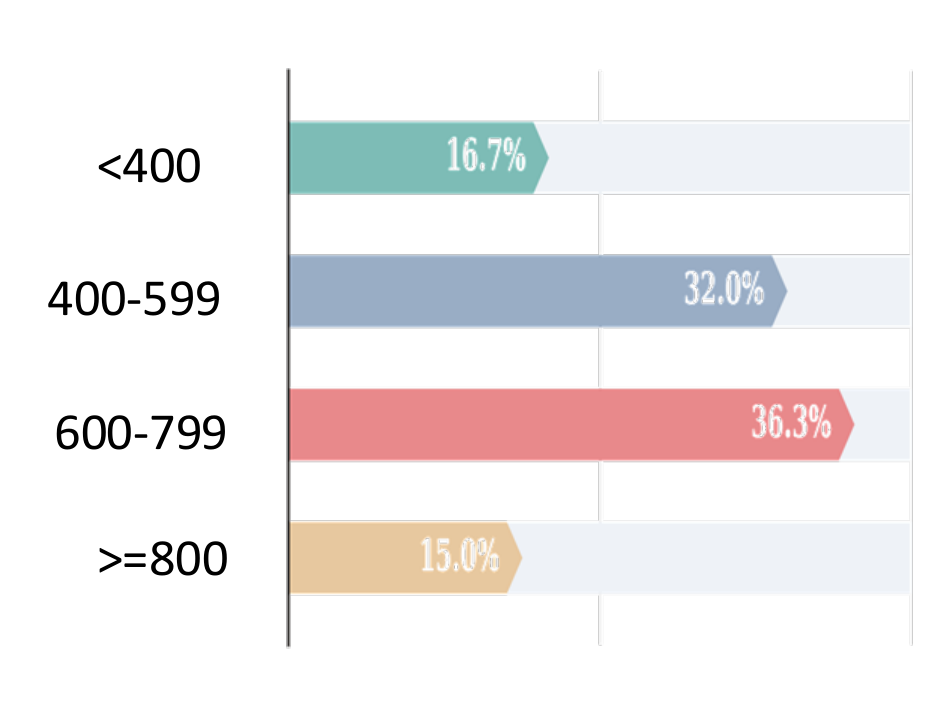}\\[-1mm]
{\footnotesize \textbf{(c)} Prompt length distribution}
\end{minipage}\hfill
\begin{minipage}[t]{0.25\textwidth}
\centering
\includegraphics[width=\linewidth]{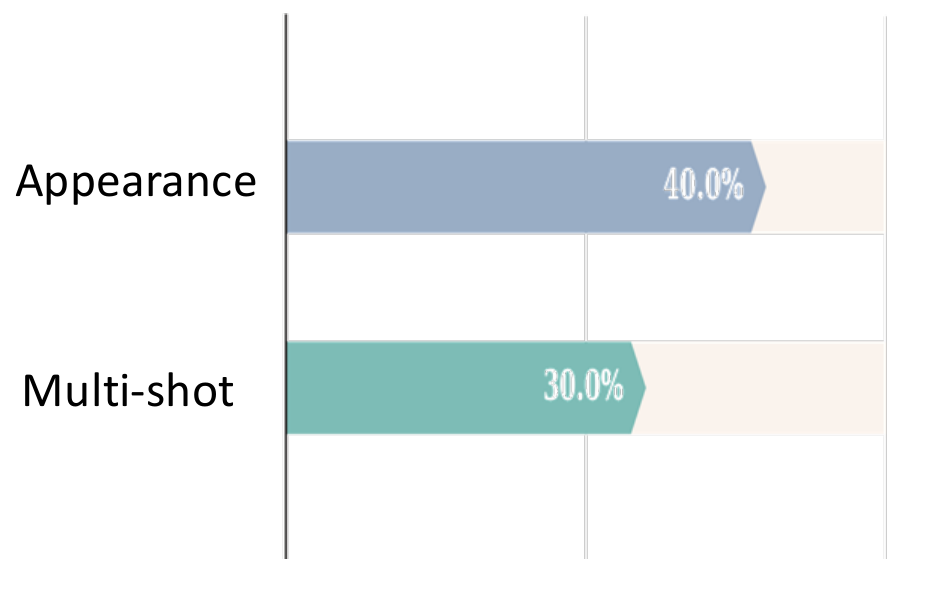}\\[-1mm]
{\footnotesize \textbf{(d)} Other requirements of prompts}
\end{minipage}

\caption{\textbf{Statistics of MultiRef-Compass.} The benchmark features complex condition combinations, including four reference structures, six audio settings, and fine-grained prompt constraints, while maintaining a balanced prompt-length distribution.}
\label{fig:dataset-statistics}
\end{figure*}
MultiRef-Compass contains $300$ samples across three main boards, supplemented by $50$ samples in the challenge board. 
Each sample is paired with a structured prompt. 
This scale balances benchmark coverage with evaluation cost, enabling practical comparisons across both proprietary and open-source models. 
As shown in Figure~\ref{fig:dataset-statistics}, the benchmark includes same-subject multi-view references and heterogeneous compositions, such as human–human–object, human–human–scene, and human–object–scene configurations. 
Overall, $70\%$ of the samples use real-world references, while the remaining $30\%$ use cartoon-style references. 
Prompts contain approximately $600$ characters on average and specify both reference usage and multimodal generation requirements. 
Because these attributes are not mutually exclusive, $60\%$ of the prompts include sound effects or ambient audio, $30\%$ include background music, $70\%$ include speech, $40\%$ impose appearance-related constraints, and $30\%$ require a multi-shot structure. 
Together, these distributions support controlled evaluation of multi-reference preservation and binding, fine-grained instruction following, and coordinated audio-visual generation.

\section{Evaluation Framework}
\subsection{Hybrid Judge Architecture}
We introduce a dual-level evaluation framework for MultiRef-Compass, as illustrated in Figure~\ref{fig:overall}.
The framework integrates automatic metrics with a rejudging-enhanced MLLM-as-a-Judge protocol~\cite{bansal2024videophy,he2024videoscore,shi2025mavors,qi2025t2veval,zhang2025omnieval}. 
For each generated sample, it evaluates videos across four major dimensions: \textit{Basic Quality}, \textit{Reference Consistency}, \textit{Audio-Visual Consistency}, and \textit{Instruction Following}. 
MLLM-based metrics are scored on a $1$–$5$ MOS scale. 
A rejudging stage is further applied to verify potentially unreliable judgments, thereby improving the robustness of the final evaluation. 
This evaluation framework follows an omni-reference design and provides fine-grained and interpretable diagnostics across diverse reference settings and modalities.

\subsection{Metric Dimensions}
Following the design specification, MultiRef-Compass evaluates each output along four metric dimensions. A
rule-based router activates only applicable metrics based on prompt metadata; for example, lip-sync checks are enabled only for dialogue or monologue samples. Automatic metrics provide objective evidence when reliable tools are available, while MLLM-based metrics use either checklist-style evaluation or direct rubric-based scoring~\cite{zhang2025ui2vbench,cao2025t2avcompass}. 
All MLLM scores are reported on a 1--5 MOS scale.

\noindent\textbf{Basic Quality (BQ).} 
We follow prior T2AV evaluation protocols and assess the basic technical quality of the generated video and audio before reference-specific evaluation.
\begin{itemize}
    \item \textbf{Visual Technical Quality (VTQ).} This metric measures low-level visual defects. We employ DOVER++~\cite{wu2023dover} to score representative frames.
    \item \textbf{Audio Technical Quality (ATQ).} This metric evaluates the basic perceptual quality and usability of the generated audio, including perceived clarity, naturalness, and the severity of audible artifacts. We use Audiobox scores~\cite{tjandra2025meta} for overall audio quality.
\end{itemize}

\begin{itemize}
    \item \textbf{Anatomical Quality (AQ).} This metric measures biological anatomical constraints, rigid-body rigidity, and scene integration with 3D depth coherence. We use an MLLM checklist to judge whether humans, objects, and backgrounds are simultaneously plausible and do not contaminate each other.
\end{itemize}

\noindent\textbf{Reference Consistency (RC).} 
We evaluate whether the generated video preserves and binds the provided references.
\begin{itemize}
    \item \textbf{Entity Fidelity (EF).}
    EF measures how faithfully generated human, object, and scene entities preserve the identity and appearance of their references. We uniformly sample video frames, using localized crops for humans and objects and full-frame comparisons for scenes. Human and object regions are detected with YOLO-World~\cite{cheng2024yoloworld} and compared with the corresponding references using SigLIP~\cite{zhai2023siglip} embeddings. For human subjects, face-specific cues are further combined with body-level appearance similarity when available. When multiple views of the same subject are provided, each generated frame is matched to its most similar reference view. Scores are then aggregated across frames and reference branches to obtain a base fidelity score $\mathrm{EF}_{\mathrm{base}}$.
    Directly copying reference content into a generated video may produce spuriously high embedding similarity without demonstrating genuine reference-conditioned synthesis. We therefore calibrate the base score using an MLLM-estimated paste-naturalness coefficient $P_{\mathrm{paste}}(r)\in[0,1]$:
    \begin{equation}
    \mathrm{EF}_{\mathrm{final}}
    =
    \mathrm{EF}_{\mathrm{base}}
    \times
    P_{\mathrm{paste}}(r).
    \end{equation}
    The coefficient preserves the similarity score when the referenced entity is naturally synthesized and discounts it when evident copy-and-paste artifacts are present. Further implementation details are provided in the Appendix~\ref{app:ef}.

    \item \textbf{Detail Preservation (DP).} This metric evaluates fine-grained detail preservation in the generated video. For background references, camera motion may prevent strict frame-level comparison, so we judge whether the regions that should remain consistent with the reference are still faithfully preserved. We evaluate it based on a MLLM.

    \item \textbf{Binding Correctness (BC).} This metric detects binding errors, including swapped identities, attributes, actions, or voices. We evaluate it based on a MLLM.
\end{itemize}

\noindent\textbf{Audio-Visual Consistency (AVC).} 
We evaluate whether sound is aligned with visible content and its source.
\begin{itemize}
    \item \textbf{Speech-Lip Synchronization (SLS).} This metric evaluates the temporal synchronization between speech and visible lip motion. To make the automatic measurement reliable, we first use an MLLM to filter for samples with stable frontal faces, and then apply LatentSync~\cite{li2024latentsync} only to dialogue or monologue samples.
    \item \textbf{Event-Sound Matching (ESM).} This metric evaluates whether visible events are accompanied by semantically appropriate sounds. We evaluate it based on a MLLM.
    \item \textbf{Source Correctness (SC).} This metric checks whether each sound comes from the correct person. We use a MLLM to verify whether each person's dialogue is emitted by the intended one and whether each character's timbre matches the specified gender.
    \item \textbf{Timbre Similarity (TS).}
    For Board 4 audio-reference inputs, this metric evaluates whether the generated speech preserves the timbre of the provided voice reference. We compute speaker-embedding similarity using SpeechBrain ECAPA-TDNN~\cite{desplanques2020ecapa}, with preprocessing and calibration details provided in the Appendix~\ref{app:ts}.
\end{itemize}

\noindent\textbf{Instruction Following (IF).} 
We evaluate whether the output follows the prompt requirements using MLLM checklists.
\begin{itemize}
    \item \textbf{Visual Task Following (VT).} This metric checks whether basic settings, important actions, wearing requirements, and details are completed. 
    \item \textbf{Audio Task Following (AT).} This metric checks whether requested sound effects and music styles are generated. 
    \item \textbf{Speech Content Accuracy (SCA).} This metric checks whether the specified spoken content is correct, fully consistent with the prompt, and in the correct language. 
    \item \textbf{Temporal Order Following (TO).} This metric checks whether event order and multi-shot sequences follow the prompt order instead of being shuffled or repeated. 
\end{itemize}

\subsection{Rejudging}
MLLM-as-a-Judge provides a flexible framework for evaluating complex reference-conditioned generation, but its judgments may vary with prompt complexity, input scale, temporal evidence coverage, and criterion ambiguity. 
When model outputs are evaluated independently, the judge may interpret the same criterion differently across samples, resulting in overly strict or permissive scores or inconsistencies between scores and rationales.

To mitigate these local inconsistencies, MultiRef-Compass applies a lightweight rejudging stage after the initial evaluation. 
For each sample and evaluation item, the rejudge jointly examines the original evaluation prompt and the initial scores and rationales assigned to all model outputs. 
Comparing these judgments side by side enables it to verify whether the same semantic criterion has been applied consistently across models. 
When the available rationales are insufficient to resolve a discrepancy, the rejudge revisits the corresponding generated videos and checks the relevant visual or temporal evidence.

The rejudge only corrects clear inconsistencies among the evaluation criterion, supporting evidence, rationale, and assigned score. 
It does not introduce new evaluation dimensions, alter metric definitions, or independently rescore otherwise consistent judgments. 
This stage therefore improves item-level consistency and auditability while preserving the original evaluation protocol.

\begin{table*}[t]
\centering
\small
\setlength{\tabcolsep}{6pt}
\begin{tabularx}{\textwidth}{@{}l c *{4}{>{\centering\arraybackslash}X}@{}}
\toprule
 & & \multicolumn{2}{c}{\textbf{Basic Quality}} & \multicolumn{1}{c}{\textbf{Reference Consistency}} & \multicolumn{1}{c}{\textbf{Audio-Visual Consistency}} \\
\cmidrule(lr){3-4}\cmidrule(lr){5-5}\cmidrule(lr){6-6}
\textbf{Model} & \textbf{Proprietary} & \textbf{VTQ} & \textbf{ATQ} & \textbf{EF} & \textbf{SLS} \\
\midrule
Seedance 2.0* & $\checkmark$ & \underline{0.2225} & \underline{7.4562} & \underline{0.6104} & \underline{3.1727} \\
Kling 3.0 & $\checkmark$ & 0.1869 & 7.3333 & \textbf{0.6160} & 2.8560 \\
HappyHouse 1.1 & $\checkmark$ & 0.1953 & \textbf{7.7729} & 0.5952 & 3.0951 \\
Gemini-Omni* & $\checkmark$ & \textbf{0.2373} & 7.3126 & 0.5967 & \textbf{4.5333} \\
Wan 2.7 & $\checkmark$ & 0.1846 & 7.2296 & 0.5656 & 2.7732 \\
Vidu Q3-Mix & $\checkmark$ & 0.2041 & 7.1909 & 0.5763 & 3.1603 \\
\midrule
SkyReels-V3 & $\times$ & 0.2062 & -- & 0.5720 & -- \\
Wan2.1-VACE & $\times$ & 0.1633 & -- & 0.5204 & -- \\
\bottomrule
\end{tabularx}
\caption{\textbf{Results obtained with automatic tools, grouped by the available evaluation dimensions.} The \textbf{best} result is shown in \textbf{bold}, and the \underline{second-best} result is \underline{underlined}. Due to content-safety filtering, Seedance 2.0* and Gemini-Omni* are evaluated on 282 and 245 samples, respectively. }
\label{tab:auto-results}
\end{table*}
\begin{table*}[ht!]
\centering
\small
\setlength{\tabcolsep}{5pt}
\resizebox{\textwidth}{!}{%
\begin{tabular}{@{}lcccccccccc@{}}
\toprule
 & & \multicolumn{1}{c}{\textbf{Basic Quality}} 
 & \multicolumn{2}{c}{\textbf{Reference Consistency}} 
 & \multicolumn{2}{c}{\textbf{Audio-Visual Consistency}} 
 & \multicolumn{4}{c}{\textbf{Instruction Following}} \\
\cmidrule(lr){3-3}\cmidrule(lr){4-5}\cmidrule(lr){6-7}\cmidrule(lr){8-11}
\textbf{Model} & \textbf{Proprietary} & \textbf{AQ} & \textbf{BC} & \textbf{DP} & \textbf{ESM} & \textbf{SC} & \textbf{VT} & \textbf{AT} & \textbf{SCA} & \textbf{TO} \\
\midrule
Seedance 2.0* & $\checkmark$ & \textbf{3.9452} & \textbf{4.6360} & \textbf{4.8500} & \textbf{4.8528} & 4.8492 & \textbf{4.7054} & \underline{4.6377} & 4.7850 & 4.7123 \\
Kling 3.0 & $\checkmark$ & 3.8994 & 4.5180 & \underline{4.8000} & \underline{4.8421} & 4.8792 & 4.4508 & \textbf{4.7101} & \textbf{4.8630} & 4.6919 \\
HappyHouse 1.1 & $\checkmark$ & 3.4909 & 4.5740 & \textbf{4.8500} & 4.8181 & \textbf{4.9441} & \underline{4.6954} & 4.1932 & \underline{4.8537} & 4.7033 \\
Gemini-Omni* & $\checkmark$ & \underline{3.9100} & \underline{4.6290} & 4.4370 & 4.8079 & \underline{4.8933} & 4.6865 & 4.6114 & 4.7233 & \underline{4.7481} \\
Wan 2.7 & $\checkmark$ & 3.7869 & 4.5110 & 4.7970 & 4.7458 & 4.8699 & 4.5319 & 4.3816 & 4.4100 & \textbf{4.7611} \\
Vidu Q3-Mix & $\checkmark$ & 3.3038 & 4.3570 & 4.3630 & 4.7461 & 4.8308 & 4.2875 & 4.5894 & 4.2830 & 4.2670 \\
\midrule
SkyReels-V3 & $\times$ & 3.3761 & 4.2410 & 4.0200 & -- & -- & 3.6126 & -- & -- & 3.0944 \\
Wan2.1-VACE & $\times$ & 3.0923 & 3.8582 & 3.4790 & -- & -- & 3.5198 & -- & -- & 3.0925 \\
\bottomrule
\end{tabular}
}
\caption{\textbf{Results obtained using MLLM-based metrics, grouped according to the four evaluation dimensions.} The \textbf{best} result is shown in \textbf{bold}, and the \underline{second-best} result is \underline{underlined}. Due to content-safety filtering, Seedance 2.0* and Gemini-Omni* are evaluated on 282 and 245 samples, respectively.}
\label{tab:mllm-results}
\end{table*}

\section{Experiments}
\subsection{Experimental Settings}

\paragraph{Evaluated Models.}
We evaluate $8$ representative models that support multi-image reference inputs for omni-modal video generation, including $6$ closed-source models and $2$ open-source models. 
The closed-source models include Seedance 2.0~\cite{seedance2026seedance}, Kling 3.0~\cite{team2025kling}, HappyHouse 1.1~\cite{happyhorse2026}, Gemini-Omni~\cite{gemini_omni_2026}, Wan 2.7~\cite{wan2025wan}, and Vidu Q3-Mix~\cite{bao2024vidu}. The open-source models include Wan2.1-VACE~\cite{jiang2025vace} and SkyReels-V3~\cite{li2026skyreelsv3}. We exclude audio metrics for open-source models as they lack native audio outputs.

\paragraph{Implementation Details.}
All experiments are conducted on eight NVIDIA A800 GPUs. 
Proprietary models are accessed through their official generation APIs when available and otherwise through their official platforms; unless constrained by the platform, they generate $10$-second videos at 1080p. 
For open-source models, we follow the officially recommended inference configurations, including duration and resolution, because enforcing a uniform output specification can substantially degrade their generation quality and depart from their intended operating settings.
To accommodate different visual-conditioning interfaces, each model receives three reference views. 
For most human-centered samples in Boards B2 and B3, these views are combined into a single stitched multi-view image when the model accepts only one reference image. 
All outputs for the same sample are evaluated using identical checklists and evaluation prompts. 
Gemini 3.1 Pro~\cite{gemini31pro} serves as the judge for all MLLM-based metrics, and repeated judging runs are conducted to assess judgment consistency and score stability.

\begin{figure*}[t]
\centering
\begin{minipage}[t]{0.245\textwidth}
\centering
\includegraphics[width=\linewidth]{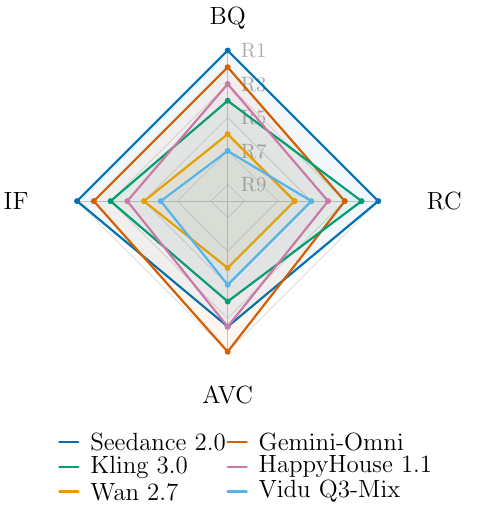}
\vspace{-1mm}
{\footnotesize \textbf{(a)} Closed-source ranking}
\end{minipage}
\hfill
\begin{minipage}[t]{0.245\textwidth}
\centering
\includegraphics[width=\linewidth]{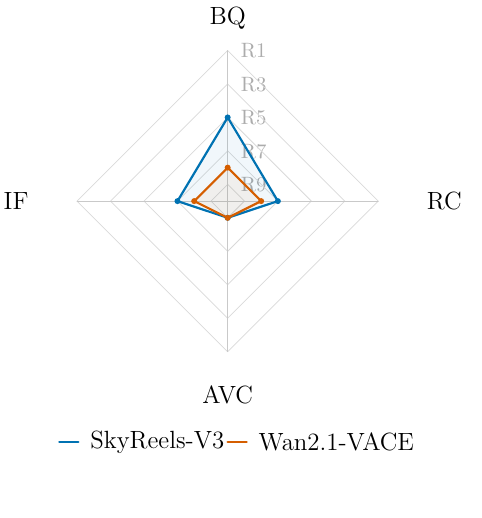}
\vspace{-1mm}
{\footnotesize \textbf{(b)} Open-source ranking}
\end{minipage}
\hfill
\begin{minipage}[t]{0.245\textwidth}
\centering
\includegraphics[width=\linewidth]{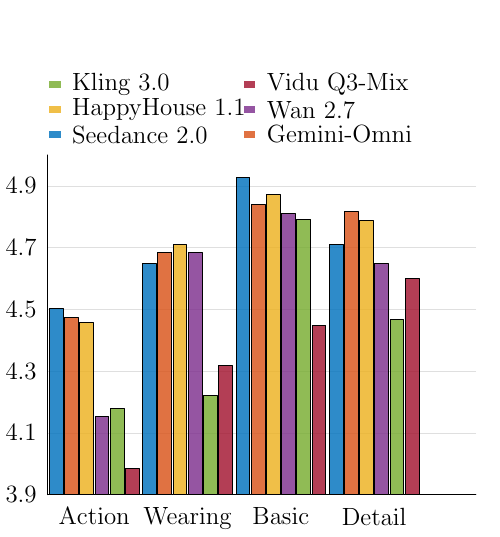}
\vspace{-1mm}
{\footnotesize \textbf{(c)} Closed-source visual IF}
\end{minipage}
\hfill
\begin{minipage}[t]{0.245\textwidth}
\centering
\includegraphics[width=\linewidth]{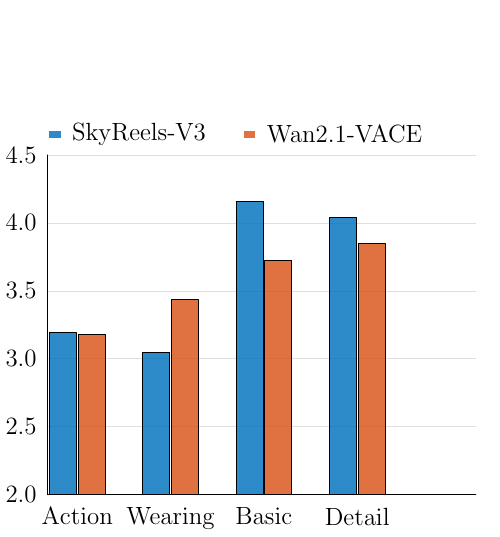}
\vspace{-1mm}
{\footnotesize \textbf{(d)} Open-source visual IF}
\end{minipage}
\caption{\textbf{Model capability visualization.} Fig.(a) and Fig.(b) show four-dimensional rankings for closed-source and open-source models. (using ranking due to different scale of auto and mllm In the radar charts, ranks range from R1 (best) at the outermost ring to R8 (worst) at the center. Fig.(c) and Fig.(d) decompose visual task following into Action, Wearing, Basic Setting, and Detail using original MLLM scores.}
\label{fig:radar-analysis}
\end{figure*}

\subsection{Main Results}
We evaluate the three main boards using 13 metrics grouped into four dimensions: \textit{Basic Quality}, \textit{Reference Consistency}, \textit{Audio-Visual Consistency}, and \textit{Instruction Following}. Tables~\ref{tab:auto-results} and~\ref{tab:mllm-results} present the automatic and MLLM-based results, respectively. Timbre Similarity is evaluated exclusively on the challenge board (B4), with the corresponding results reported in Appendix~\ref{app:challenge-board}.

\paragraph{Basic Quality.}
Gemini-Omni and HappyHouse 1.1 lead in visual and audio technical quality, respectively, while Seedance 2.0 achieves strong structural realism and therefore obtains a high AQ score. Notably, Seedance 2.0 shows occasional face-region artifacts in real-person cases, likely due to stricter safety processing. Overall, the results suggest a clear trade-off among visual, clarity, and anatomical quality in current MR2AV systems.

\begin{figure}[t]
\centering
\begin{minipage}[t]{0.49\linewidth}
\centering
\includegraphics[width=\linewidth]{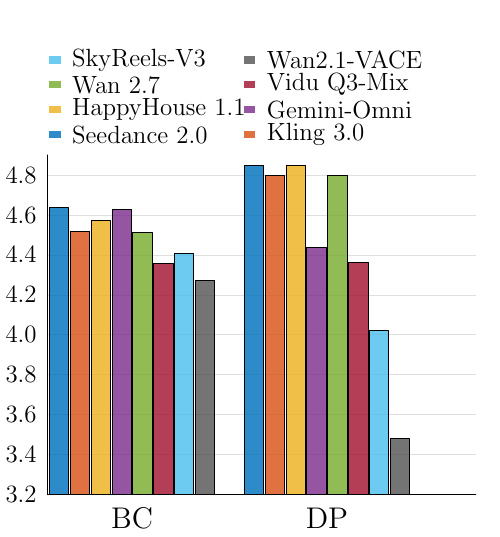}
\vspace{-1mm}
{\footnotesize \textbf{(a)} BC / DP}
\end{minipage}
\hfill
\begin{minipage}[t]{0.49\linewidth}
\centering
\includegraphics[width=\linewidth]{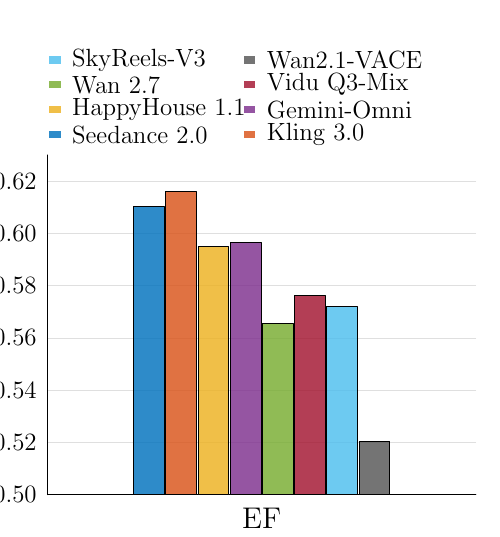}
\vspace{-1mm}
{\footnotesize \textbf{(b)} EF}
\end{minipage}
\caption{\textbf{Reference consistency breakdown across all evaluated models.} BC and DP are reported with MLLM-based scores, while EF reports automatic entity fidelity.}
\label{fig:rc}
\end{figure}


\paragraph{Reference Consistency.}
\begin{figure}[h]
\centering
\includegraphics[width=.48\textwidth,height=.31\textheight]
{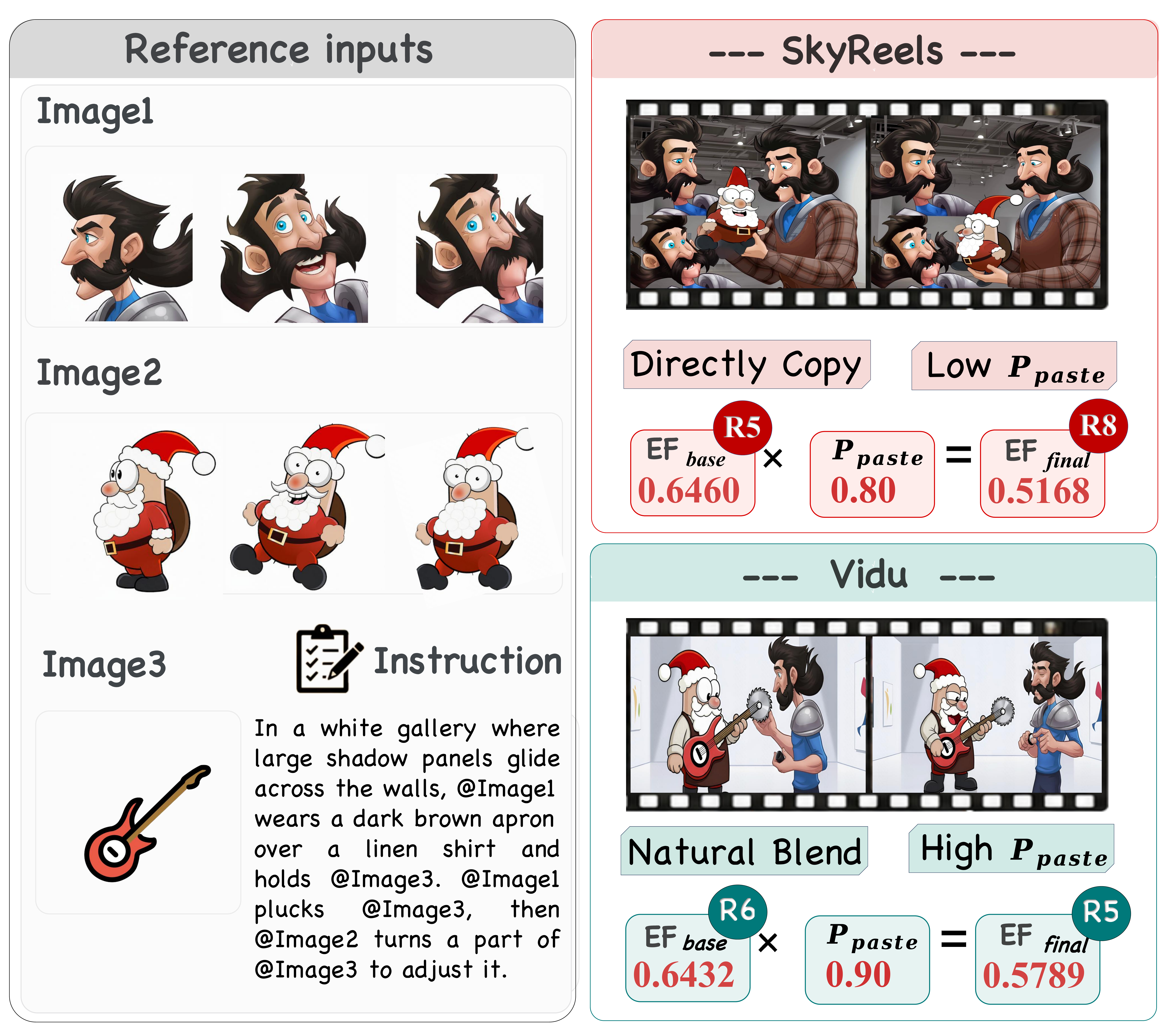}
\caption{\textbf{Representative SkyReels-V3 failure case showing copy-and-paste artifacts and identity splitting.} After incorporating the paste-naturalness factor $P_{\mathrm{paste}}(\cdot)$, SkyReels-V3 drops from rank 5 to rank 8 because of its strong pasted appearance, whereas Vidu Q3-mix rises from rank 6 to rank 5 thanks to more natural visual integration.}
\label{fig:panelty}
\end{figure}

Kling 3.0 achieves the highest EF, indicating stronger overall visual similarity to the provided references. 
In contrast, Seedance 2.0 performs best on BC and DP, showing stronger binding and detail retention. 
Figure~\ref{fig:rc} provides a more fine-grained view of the three Reference Consistency metrics. 
This suggests that EF and MLLM-based reference judgments capture complementary aspects: high embedding-based similarity does not always imply the best fine-grained preservation. 
Moreover, we observe that current models, particularly open-source ones, still struggle with both natural visual integration and cross-reference understanding. 
As shown in Figure~\ref{fig:panelty}, SkyReels-V3 directly copies a reference image into the generated video, producing an unnatural copy-and-paste artifact that nevertheless receives a high similarity score from conventional automatic metrics. 
Our $P_{\mathrm{paste}}$ discounts this score inflation and yields a ranking that more faithfully reflects actual generation ability. 
The model also misinterprets multiple views of the same subject as distinct identities, resulting in identity splitting.
Beyond these failures, HappyHouse 1.1 often produces polished outputs, but with a noticeably more synthetic appearance. 
Additional qualitative observations further suggest that Veo exhibits weaker identity preservation for several Asian face references in our test set, whereas the later Gemini-Omni release shows improved preservation in these cases. 
More representative cases are provided in the Appendix.

\paragraph{Audio-Visual Consistency.}
Gemini-Omni obtains the highest SLS, indicating the strongest lip synchronization among the evaluated models. 
Seedance 2.0 leads ESM, while HappyHouse 1.1 leads SC, indicating that semantic event-audio matching and correct sound-source attribution remain distinct challenges.

\paragraph{Instruction Following.}
The results reveal distinct strengths across visual, audio, speech, and temporal compliance. 
Seedance 2.0 shows its main advantage in the visual task, Kling 3.0 is stronger in audio and speech-related instruction compliance, and Wan 2.7 performs well in the temporal task. 
This indicates that current models are not uniformly strong across all instruction types, but instead exhibit modality-specific advantages.

\subsection{Further Analysis}
Figure~\ref{fig:radar-analysis} (a) and Figure~\ref{fig:radar-analysis} (b) present the overall rankings and four-dimensional capability profiles of the evaluated models, computed on the shared subset of videos successfully generated by all systems. 
Among the closed-source models, Seedance 2.0 ranks first overall, owing to its consistently strong performance across all four evaluation dimensions. 
Gemini-Omni ranks second, with notable advantages in Basic Quality and Audio-Visual Consistency, while Kling 3.0 and HappyHouse 1.1 follow closely with complementary strengths in Reference Consistency and Instruction Following. 
Wan 2.7 and Vidu Q3-Mix rank lower overall, although each remains competitive on selected sub-metrics. 
These results show that the leading position of Seedance 2.0 arises from its relatively balanced capability profile, whereas the other systems tend to exhibit more specialized strengths.

In contrast, the capability profiles of the open-source models are consistently compressed toward the inner rings. 
SkyReels-V3 performs better overall than Wan2.1-VACE, but both remain substantially behind the closed-source systems across the evaluated dimensions. 
In particular, their weaker performance in Reference Consistency and Instruction Following indicates that current open-source models still struggle not only to preserve individual references, but also to understand relationships among multiple references and bind them correctly to prompt-specified entities and events.

Figure~\ref{fig:radar-analysis} (c) and Figure~\ref{fig:radar-analysis} (d) further decompose Visual Task Following into Action, Wearing, Basic Setting, and Detail. 
Among closed-source models, performance on Basic Setting and Detail is relatively stable, whereas Action and Wearing produce larger separations across systems. 
This suggests that current models are generally more reliable at preserving static scene attributes than at executing fine-grained subject actions or satisfying clothing-related constraints. 
The same pattern is more pronounced for the open-source models: both SkyReels-V3 and Wan2.1-VACE exhibit lower performance across all four sub-dimensions, with particularly incomplete fulfillment of Action and Wearing requirements. 
These findings demonstrate that Visual Task Following should not be treated as a single scalar capability, since a high aggregate VT score may conceal substantial failures in specific controllable attributes.

\paragraph{Overall Findings and Implications.}
The results reveal substantial variation across models and evaluation dimensions, with no evaluated system achieving consistently strong performance across all four dimensions. 
Reference Consistency declines as reference compositions become more complex, showing that preserving and correctly binding multiple subjects remains difficult in multi-entity settings. 
Anatomical Quality is also limited across models, with failures including implausible body structures and incoherent interactions. 
In Audio-Visual Consistency, speech is sometimes generated without corresponding lip motion, while Instruction Following degrades under complex and fine-grained constraints. 
Proprietary systems generally outperform open-source models, with the largest gaps observed in Reference Consistency and Instruction Following.

These results show that MR2AV generation requires the joint optimization of perceptual quality, structural plausibility, multi-reference coherence, and audio-visual instruction adherence. 
By evaluating these capabilities separately, MultiRef-Compass exposes capability-specific failures that can be obscured by aggregate quality scores. 
Future progress therefore depends not only on improving individual components, but also on coordinating references, entities, modalities, and temporal dynamics within a unified generation process. 
The benchmark can further accommodate richer omni-reference settings; Appendix~\ref{app:challenge-board} reports Board 4 results with audio and video references.

\paragraph{Effect of Rejudging.}
We further analyze representative cases to clarify the role of rejudging as a calibration step in MLLM-based evaluation. 
One example concerns the checklist question \emph{``Does someone adjust the position of @Image2?''}. 
Some outputs are assigned low scores because the judge interpreted ``adjust'' narrowly as referring to an explicit fine-grained adjustment gesture. 
In these videos, however, the subject picks up and places the referenced plant in a new location, which changes its position and satisfies the intended action. 
The horizontal rejudge resolves this inconsistency by applying a consistent semantic criterion across models for the same question.
As shown in the Appendix~\ref{app:rejudge}, it corrects both overly strict and overly permissive judgments, improving score consistency without altering the evaluation dimensions or the overall model-level conclusions.
 
\subsection{Human Preference Alignment}
We examine whether our evaluation metrics produce model preferences consistent with human judgments. 
For each evaluated dimension, four human annotators independently compare model outputs generated from the same input condition. 
Each comparison is encoded as a preference outcome, where the preferred output receives $1$, a tie receives $0.5$, and the non-preferred output receives $0$. 
We apply the same pairwise conversion to benchmark scores and aggregate the outcomes into model-level win rates.
As reported in Table~\ref{tab:human-alignment}, our benchmark achieves strong correlations with human preferences across all dimensions, with $0.8979$ for BQ, $0.9380$ for RC, $0.9217$ for AC, and $0.9644$ for IF. 
The results indicate that our evaluation protocol can reflect human preferences, providing an effective reference for diagnosing MR2AV models.

\begin{table}[t]
\centering
\small
\begin{tabular}{lc}
\toprule
\textbf{Dimension} & \textbf{Pearson} $\uparrow$ \\
\midrule
Basic Quality (BQ) & 0.8979 \\
Reference Consistency (RC) & 0.9380 \\
Audio-Visual Consistency (AVC) & 0.9217 \\
Instruction Following (IF) & 0.9644 \\
\bottomrule
\end{tabular}
\caption{Human preference alignment via Pearson correlation between human and benchmark win rates.}
\label{tab:human-alignment}
\vspace{-1.5mm}
\end{table}

\section{Limitations}
MultiRef-Compass is designed as a controlled diagnostic benchmark and therefore does not cover every creative domain, cultural style, or reference modality. 
To ensure comparability across models with different conditioning interfaces, the main evaluation standardizes each sample to three reference images; substantially larger reference sets remain outside the current scope.
The automatic evidence tools also have limitations, particularly for cartoon subjects and faces undergoing large pose or expression changes. 
Although audio evaluation is a central component of MultiRef-Compass, our speech–lip synchronization (SLS) metric remains sensitive to facial stability. 
As discussed in Appendix~\ref{app:sls}, models that generate stable, frontal speaking faces, such as Gemini-Omni in our experiments, may receive more favorable SLS scores, whereas models producing greater facial and head motion may be penalized even when their outputs appear natural. 
More robust synchronization metrics are therefore needed for dynamic faces and complex multi-speaker scenes.
Finally, MLLM-based judging incurs additional cost and may inherit model-specific biases. 
Because proprietary systems and their APIs may change over time, we report model versions and access dates to support reproducibility.

\section{Conclusion}
In this paper, we present MultiRef-Compass, a comprehensive benchmark for evaluating multi-reference-to-audio-video (MR2AV) generation. 
The benchmark introduces an asset-reuse pipeline for constructing diverse multi-reference evaluation samples and combines automatic metrics with a rejudging-enhanced MLLM-as-a-Judge protocol. 
It evaluates MR2AV systems along four dimensions: \textit{Basic Quality}, \textit{Reference Consistency}, \textit{Audio-Visual Consistency}, and \textit{Instruction Following}. 
Experiments on representative MR2AV systems demonstrate that MultiRef-Compass distinguishes model capabilities and exposes failure modes not captured by existing benchmarks. 
We hope it provides a practical diagnostic testbed for evaluating future omni-reference audio-video generation systems.

\bibliography{main}

\clearpage
\onecolumn
\appendix

\begin{center}
{\LARGE\bfseries Appendix}
\end{center}
\setcounter{secnumdepth}{2}
\input{appendix}
\end{document}

%% file: Appendix.tex
\section{MR2AV Definition}

Multi-reference-to-audio-video (MR2AV) generation synthesizes new audio-video content from multiple references and a textual instruction. The references are not limited to images, but may include multimodal inputs, e.g., videos and audio clips, that provide information to be preserved or reflected in the generated content. Here, ``multi-reference'' means that the model must jointly interpret and compose information from multiple reference sources. These references may depict different views of the same entity or provide distinct subjects, objects, scenes, motions, and sounds, while the textual instruction specifies their intended roles, relationships, and interactions in the generated event.

MR2AV is a reference-conditioned generation task rather than an editing task. A reference video may provide subject identity, appearance, or motion cues, while a reference audio clip may provide voice identity, timbre, or acoustic characteristics. Neither is treated as source content to be directly edited.

\section{Dataset Construction Details}

This section provides additional details on the construction of MultiRef-Compass, including asset-pack collection, board construction, the motivation behind the board design, and the structured prompt generation procedure. The goal of the construction pipeline is to create a benchmark that is realistic enough to reflect practical multi-reference R2AV workflows, while remaining controlled enough to support reproducible and diagnostic evaluation.
\subsection{Overview and Design Rationale}

We adopt an asset-pack construction strategy to make MultiRef-Compass both controlled and diagnostic. Instead of collecting each benchmark sample as an isolated prompt, we first build reusable subject, object, scene, video, and audio packs, and then recombine them according to board-specific rules. This design reduces redundant annotation, since each pack only needs to be verified once for quality, identity consistency, and usage constraints. It also makes the source of difficulty more interpretable: Board 1 focuses on multi-view subject preservation, Board 2 evaluates heterogeneous reference composition, and Board 3 stresses multi-entity role binding. These settings reflect common user workflows, such as providing several images of the same character, combining a person with an object in a target scene, or asking multiple referenced entities to interact. At the model interface, each sample contains \textbf{three reference-input slots}, following the input limit of several evaluated models. Some subject references combine multiple views into a single input; when these views are counted separately, each sample contains \textbf{three to seven atomic visual inputs}. This design keeps the model-level input format fixed while varying the number and relationships of the underlying references. Moreover, because each sample records which references are present and what role each reference plays, the evaluation framework can activate the corresponding metrics, enabling fine-grained diagnosis of reference fidelity, audio-visual consistency, and instruction following.
\subsection{Asset-Pack Collection}

\paragraph{Subject packs.}
Subject packs provide the primary identity references. Each subject pack contains multiple images of the same person or character. We construct subject packs through two complementary paths. For real-world subjects, we sample frames from permissively licensed or copyright-free videos and manually select frames with consistent identity and clear face/body visibility from Pexels. When only one high-quality canonical image is available, we use Nano Banana to synthesize auxiliary references with different viewpoints, expressions, poses, or speaking states. To test whether models can distinguish the target subject from the surrounding context, we generate these multi-view references with diverse and clearly visible backgrounds, preventing the benchmark from relying on a fixed background as a shortcut for identity preservation.

\paragraph{Object and Scene Packs.}
Object and scene packs provide non-subject references for constructing heterogeneous multi-image compositions. They are all collected from Pexels. Object packs include visually distinguishable foreground entities, such as daily objects, clothing, accessories, tools, food, toys, and small animals, which can appear as wearable items, handheld tools, scene props, or interaction targets. Scene packs provide environmental references from natural, urban, indoor, and stylized settings. We retain references with clear visual identity, recognizable structure, and plausible compatibility with other packs, while filtering out ambiguous, heavily occluded, or overly cluttered cases. These packs allow the benchmark to evaluate whether a model can preserve non-human references and place them into the correct roles, environments, and interactions, rather than merely generating a generic object category or background.

\paragraph{Audio Packs.}
Audio packs provide voice references for matching different subject types. We collect a diverse set of timbres, including young female, elderly female, young male, elderly male, high-pitched cartoon, and low-pitched cartoon voices. Realistic human voices are collected from permissively licensed sources such as Freesound, while cartoon-style voices are synthesized using a speech generation model. These audio packs are used to construct speech-conditioned samples and to evaluate whether models can generate speech that is consistent with the intended speaker type.

\paragraph{Video Packs.}
Video packs provide dynamic subject references. For real-world subject packs constructed from videos, we retain the original source clips as video references when available. These video packs are provided as additional subject references, allowing us to evaluate whether R2AV systems can follow the prompt-specified content while using the video only as reference information, rather than directly copying its original motion or events.

\paragraph{Metadata-Guided Pack Matching.}

Each asset pack is accompanied by a JSON metadata file that records its category, style, key attributes, and applicable usage constraints. These metadata files support automatic pack matching during sample construction and help avoid incompatible combinations, such as pairing cartoon-style objects with realistic human subjects.
\begin{figure}[t]
\centering
\includegraphics[width=0.85\linewidth]{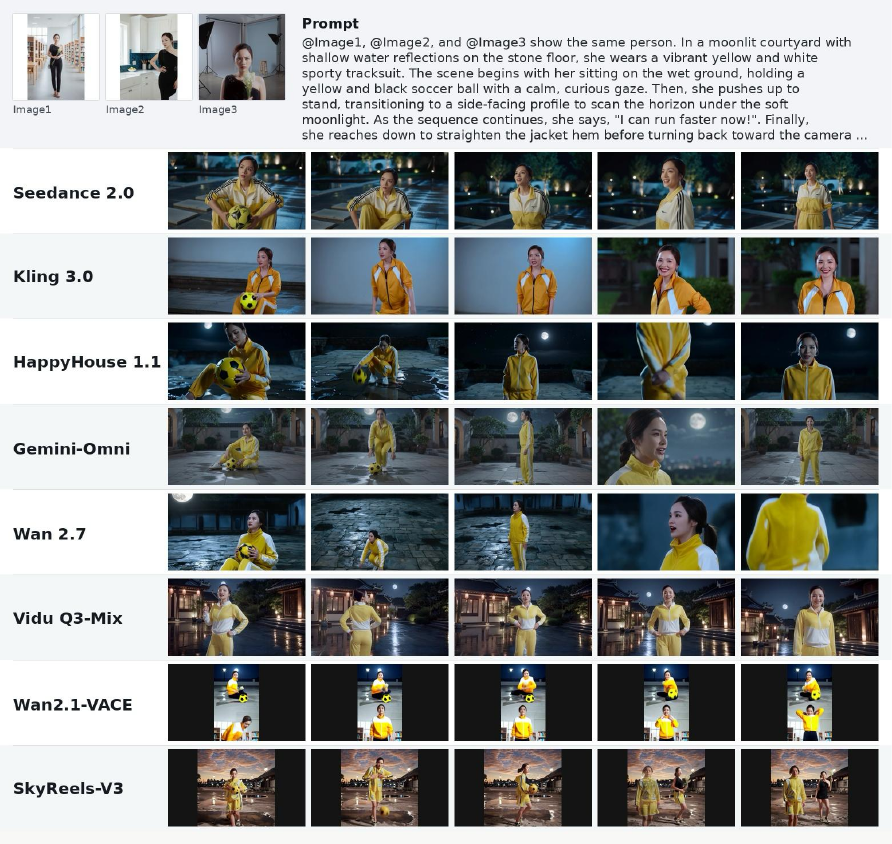}
\caption{Representative example of Board 1.}
\label{fig:board1-example}
\end{figure}

\begin{figure}[t]
\centering
\includegraphics[width=0.85\linewidth]{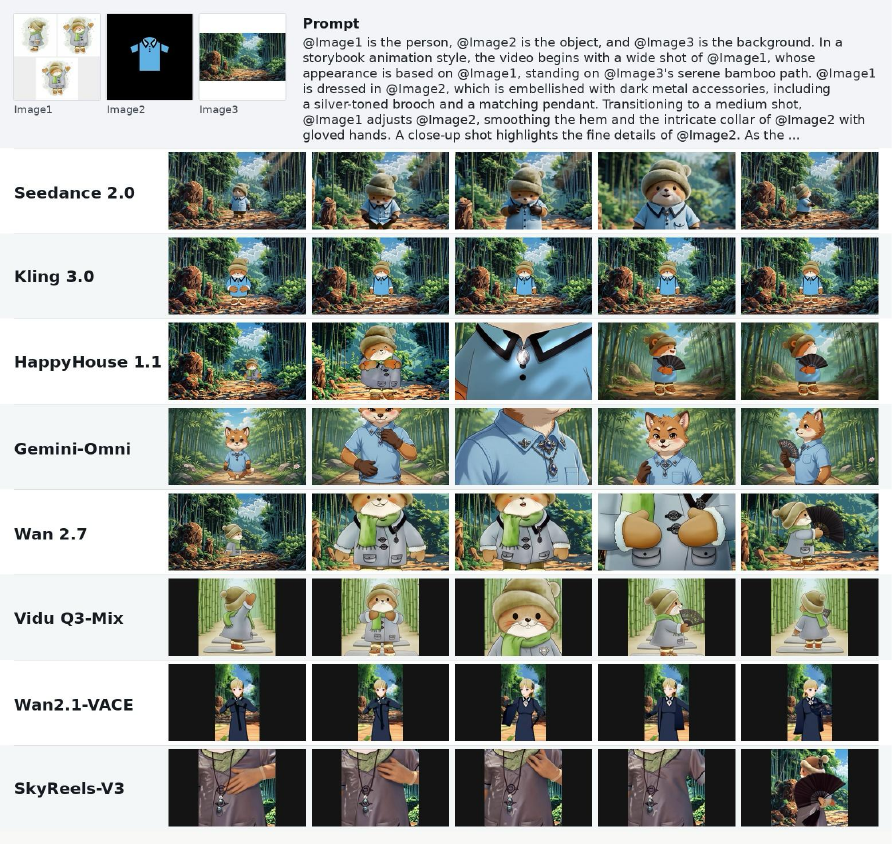}
\caption{Representative example of Board 2.}
\label{fig:board2-example}
\end{figure}

\begin{figure}[t]
\centering
\includegraphics[width=0.85\linewidth]{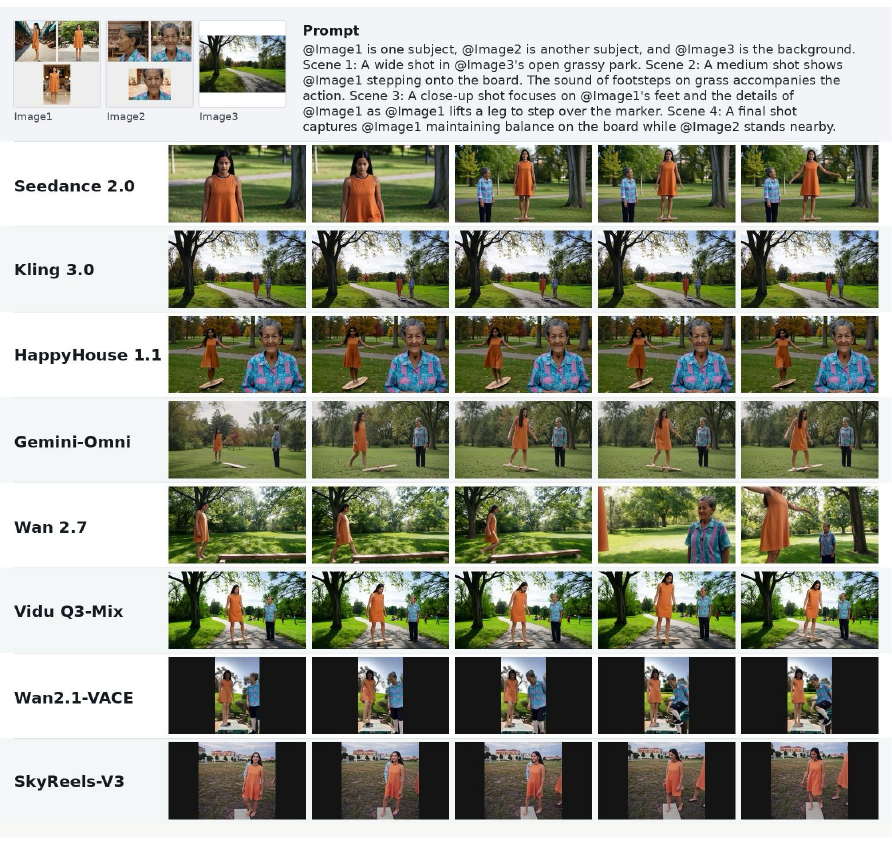}
\caption{Representative example of Board 3.}
\label{fig:board3-example}
\end{figure}

\subsection{Board Construction}

\paragraph{Board 1: Multi-view Subject Preservation.}
As shown in figure~\ref{fig:board1-example}, board 1 uses multiple images from the same subject pack, including different viewpoints, expressions, poses, or speaking states. The goal is to test whether a model can synthesize these references into one coherent dynamic subject while preserving identity and appearance over time. This board also checks whether the model mistakenly generates multiple subjects from different views of the same identity.

\paragraph{Board 2: Heterogeneous Reference Composition.}
As shown in figure~\ref{fig:board2-example}, board 2 combines subject, object, and scene-related references in flexible forms, such as subject-object-scene mixtures, multi-view subject with scene references, or multiple subjects grounded in a shared scene. Unlike Board 3, this board focuses on heterogeneous composition across reference types rather than explicit entity-role binding. It evaluates whether a model can preserve different kinds of references and place them into a coherent visual context, instead of ignoring the object, changing the scene, or letting one reference type dominate the others.

\paragraph{Board 3: Multi-entity Binding.}
As shown in figure~\ref{fig:board3-example}, board 3 samples multiple entities from different subject or object packs and assigns them distinct roles in a shared event. The prompt may specify who performs an action, who receives or manipulates an object, which entity should speak, or where each entity should appear. This board directly tests reference binding: a model may preserve each entity individually but still fail by swapping identities, merging entities, transferring attributes, or assigning the correct action to the wrong target.

\paragraph{Board 4: Audio-Video Reference Input Challenge.}
Since only a limited number of models currently support video or voice references and outputs, source-video and voice-reference packs are used in additional extension boards rather than the core benchmark. These boards allow us to evaluate whether models can use dynamic or timbre references while still following the prompt-specified content, without changing the main board construction logic.

\subsection{Prompt Schema and Sample Records}

\begin{table}[t]
\centering

\begin{tabular}{p{2.7cm}p{15cm}}
\toprule
\textbf{Slot Type} & \textbf{Description} \\
\midrule
Reference slots &
Identifiers for subject, object, scene, and optional source-video or voice references. \\
\midrule
Visual task slots &
Visual requirements such as overall style, subject appearance, clothing, scene setting, action, state, expression, and object interaction. \\
\midrule
Audio slots &
Environmental sounds, object-related sounds, and background music. \\
\midrule
Speech slots &
Speaker identity, language, utterance content, and dialogue order. \\
\midrule
Temporal slots &
Single-shot or multi-shot structure, action order, state changes, and final pose or scene state. \\
\bottomrule
\end{tabular}
\caption{Structured prompt schema of MultiRef-Compass, with fields activated by board and sample requirements.}
\label{tab:prompt_schema}
\end{table}

For each sampled asset composition, we generate prompts through a structured schema rather than free-form writing. Depending on the board setting and sample difficulty, a prompt may include fields for visual grounding, action and temporal structure, audio events, speech content, and the final model-facing instruction. Not all fields are required for every sample; instead, they are included according to the overall sample complexity. The prompt structure is summarized in Table~\ref{tab:prompt_schema}. 
We then use an LLM-assisted drafting process to instantiate prompts from the schema. Given the composed asset packs, their metadata, and the target difficulty level, the LLM drafts four fields: a visual/video prompt, an audio prompt, a speech prompt, and a final integrated prompt. The final prompt is manually reviewed and refined to remove unreasonable requirements, inconsistent reference usage, and unsafe content such as violent or sexual scenarios. This design makes MultiRef-Compass more suitable for fine-grained testing, since each evaluation dimension can be applied according to the specific requirement.

\section{Board4 Analysis}

\begin{table}[t]
\centering
\small

\label{tab:additional-leaderboard}
\resizebox{\textwidth}{!}{%
\begin{tabular}{@{}lcccccc@{}}
\toprule
\multicolumn{7}{c}{\textit{\textbf{Automatic Metrics}}} \\
\midrule
 & & \multicolumn{2}{c}{\textbf{Basic Quality}} & \multicolumn{1}{c}{\textbf{Reference Consistency}} & \multicolumn{2}{c}{\textbf{Audio-Visual Consistency}} \\
\cmidrule(lr){3-4}\cmidrule(lr){5-5}\cmidrule(lr){6-7}
Model & Closed-source & VTQ & ATQ & EF & SLS & TS \\
\midrule
Seedance 2.0 & Yes & \textbf{0.200} & 7.129 & \textbf{0.614} & \textbf{2.690} & \textbf{2.690} \\
Kling 3.0 & Yes & 0.185 & \textbf{7.226} & 0.606 & 2.324 & 1.724 \\
\bottomrule
\end{tabular}
}

\vspace{2mm}

\resizebox{\textwidth}{!}{%
\begin{tabular}{@{}lccccccccc@{}}
\toprule
\multicolumn{10}{c}{\textit{\textbf{MLLM-based Metrics}}} \\
\midrule
 & \multicolumn{1}{c}{\textbf{Basic Quality}} & \multicolumn{2}{c}{\textbf{Reference Consistency}} & \multicolumn{2}{c}{\textbf{Audio-Visual Consistency}} & \multicolumn{4}{c}{\textbf{Instruction Following}} \\
\cmidrule(lr){2-2}\cmidrule(lr){3-4}\cmidrule(lr){5-6}\cmidrule(lr){7-10}
Model & AQ & BC & DP & ESM & SC & VT & AT & SCA & TO \\
\midrule
Seedance 2.0 & \textbf{3.618} & \textbf{4.710} & \textbf{4.924} & \textbf{4.842} & 4.857 & \textbf{4.170} & \textbf{4.484} & 4.000 & 4.683 \\
Kling 3.0 & 3.148 & 4.620 & 4.679 & 4.708 & \textbf{5.000} & 4.122 & 4.448 & \textbf{4.563} & \textbf{4.750} \\
\bottomrule
\end{tabular}
}
\caption{Additional Board 4 leaderboard for selected models. The metric columns follow the automatic and MLLM-based result tables in the main paper. TS denotes voice timbre similarity between the generated speech and the provided audio reference.}
\label{app:challenge-board}
\end{table}
Board 4 follows the core sample types of Board 2 and Board 3, but adds audio-video reference inputs to create more complex multimodal interactions.
Board 4 largely follows the trends of Boards 1--3, but introduces additional challenges from audio-video reference conditioning. In \textbf{Basic Quality}, Kling 3.0 obtains slightly higher audio technical quality, while Seedance 2.0 achieves higher visual technical quality and a clearly stronger anatomical quality score, suggesting better structural stability under this challenging subset. In \textbf{Reference Consistency}, Seedance 2.0 leads on both automated EF and the MLLM-based binding and detail-preservation metrics, indicating more reliable preservation of referenced identities and fine-grained visual attributes.
In \textbf{Audio-Visual Consistency}, Seedance 2.0 performs better on event-sound matching, speech-lip synchronization, and timbre similarity, whereas Kling 3.0 obtains the higher source-correctness score. The TS results show that neither model reliably preserves the input voice timbre in Board 4, but Seedance 2.0 achieves a relatively higher score, indicating closer timbre similarity to the provided audio reference. This suggests that current models still struggle to use voice references faithfully, even when they can generate plausible speech or audio events.
In \textbf{Instruction Following}, the results are mixed: Seedance 2.0 is slightly stronger on visual and audio task following, while Kling 3.0 performs better on speech content accuracy and temporal order following. 

Overall, Board 4 preserves the same broad pattern as the main benchmark: Seedance 2.0 is more balanced across reference preservation, structural quality, and audio-reference use, while Kling 3.0 remains competitive on selected instruction and source-attribution dimensions.

\section{Case Analyses}

This section provides additional case-level and board-level analyses for representative metrics. These analyses complement the metric construction details below by showing how the evaluation protocol behaves under different board settings and how rejudging corrects inconsistent checklist judgments.

\subsection{Entity Fidelity Analysis}

We further analyze Entity Fidelity (EF) across the three boards to understand how reference complexity affects visual preservation. Board 1 obtains the highest average EF score of 0.6930, while Board 2 and Board 3 drop to 0.5893 and 0.5748, respectively. Branch-level results show that this decline is mainly driven by the Human branch: average Human EF decreases from 0.6930 on Board 1 to 0.5522 on Board 2 and 0.5409 on Board 3, whereas Object and Background scores remain relatively stable around 0.65--0.68. This indicates that the harder boards primarily stress human identity, face/body appearance, and temporal consistency under heterogeneous composition, interaction, occlusion, and pose variation, rather than simply making object or background preservation harder. Meanwhile, this drop may also be affected by metric sensitivity: multi-person interactions are more likely to produce profile faces, smaller faces, or partial occlusions, which may introduce a downward bias in face/body similarity scores.

At the model level, stronger systems such as Kling and Seedance remain relatively stable across boards and rank among the top models on all three board totals. Other models exhibit different branch-level weaknesses: some are more limited by human consistency, while others are affected by object or background preservation. Overall, these patterns suggest that board-level EF differences reflect both increased task difficulty and model-specific robustness. As multi-reference composition and entity interaction become more complex, maintaining human consistency becomes the dominant challenge for current R2AV systems.

\begin{figure}[t]
\centering
\includegraphics[width=1.0\textwidth]{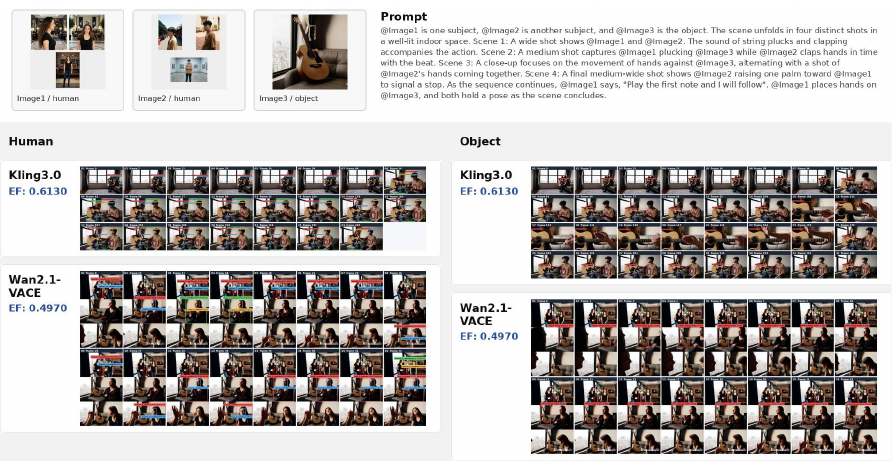}
\caption{Representative Entity Fidelity case. This case illustrates the operation of EF tools.}
\label{fig:ef-case-examples}
\end{figure}

\subsection{Speech-Lip Synchronization Analysis}
\label{app:sls}
The pre-filter removes videos with large head motion, invisible speaking faces, off-screen speech, or insufficient mouth visibility, but residual motion can still affect SyncNet-based scores. The retained sample count also differs across models: Gemini-Omni, which often produces stable frontal faces, retains 84 valid videos, while more dynamic models such as Kling and Seedance 2.0 retain only 47 valid videos. These results suggest that future lip-sync evaluation should develop more robust synchronization models that can tolerate dynamic motion, profile faces, occlusions, and other challenging speaking conditions.

The board-level results suggest that speech-lip synchronization evaluation becomes less stable as the generation setting becomes more complex. Board 1 retains the largest number of valid samples after the stable-frontal and temporal-offset filtering, indicating that it contains more visible and stable speaking-face cases. In contrast, Board 2 and Board 3 retain substantially fewer valid samples across models, making their VSLS scores more sensitive to sample selection. Although Board 2 can sometimes achieve high scores on the retained videos, these scores should be interpreted cautiously due to the limited number of valid samples. Board 3 is generally the most challenging setting, with the lowest retention counts and more cases affected by multi-entity interactions, pose variation, and reduced mouth visibility. Therefore, VSLS should be analyzed together with board type and retained sample count, rather than treated as a purely model-intrinsic synchronization score.

\begin{figure}[t]
\centering
\includegraphics[width=\textwidth]{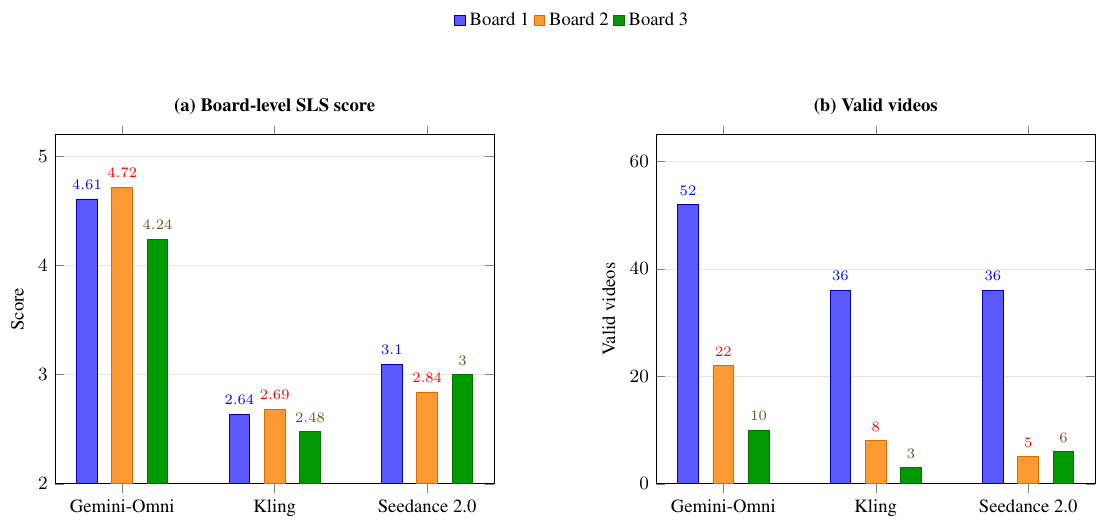}
\caption{Board-level Speech-Lip Synchronization analysis. (a) SLS scores across the three boards for Gemini-Omni, Kling, and Seedance 2.0, grouped by model. (b) Number of valid videos retained after filtering.}
\label{fig:sls-board-analysis}
\end{figure}

\paragraph{Speech Content Accuracy Analysis.}
We analyze Speech Content Accuracy (SCA) according to the requested speech language. At the aggregate level, English and Chinese achieve nearly identical strict accuracy, with 87.8\% for English and 88.0\% for Chinese, suggesting that the benchmark does not strongly favor either language overall. However, model-level results reveal clear language-specific behaviors.

Kling achieves 90.6\% accuracy on English and 97.6\% on Chinese, while HappyHouse 1.1 reaches 89.9\% on English and 100.0\% on Chinese. Their remaining errors are mainly mild English-side failures, including extra words, repeated phrases, natural paraphrases, and occasional omission of required English content. Seedance 2.0 also performs strongly in both languages, with 93.5\% English accuracy and 91.2\% Chinese accuracy, but exhibits occasional language-control and content-control errors. Some Chinese prompts are rendered in English or another foreign language, some English prompts are generated in Chinese, and some outputs introduce additional speech content beyond the required utterance. Gemini-Omni shows the strongest English performance, reaching 98.1\%, but its Chinese accuracy drops to 76.7\%. Its Chinese-side failures are mainly caused by word substitutions, semantic drift, missing speech, unintelligible speech, and a small number of unrelated English outputs. In contrast, Wan and Vidu are less stable across both languages. Wan obtains 82.6\% English accuracy and 84.0\% Chinese accuracy, while Vidu obtains 74.1\% and 75.0\%, respectively. Their errors are dominated by missing speech, irrelevant utterances, and occasional translation of Chinese requirements into English. Gemini-Omni has more Chinese-side errors, including additions, repetitions, unrelated dialogue, English substitution for Chinese prompts, and semantic reversal.

These results indicate that SCA measures more than whether a model can synthesize speech. It also captures whether the model preserves the requested language, lexical content, dialogue assignment, and prompt-specified utterance structure. This motivates evaluating SCA separately from speech-lip synchronization and audio quality: a model may generate fluent and synchronized speech while still failing to reproduce the required spoken content.

\begin{figure}[t]
\centering
\makebox[\textwidth][c]{%
\includegraphics[width=0.85\textwidth]{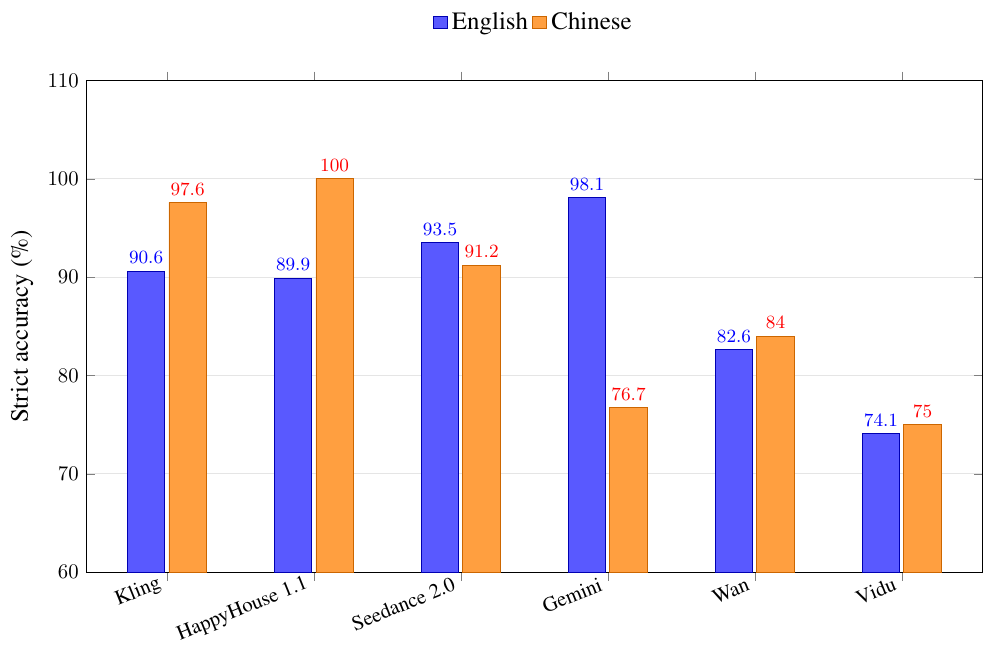}
}
\caption{Language-dependent performance in Speech Content Accuracy. The figure reports strict accuracy on English and Chinese speech items for each model.}
\label{fig:sca-language-analysis}
\end{figure}
\subsection{Rejudging Cases Analysis}
\label{app:rejudge}
Table~\ref{tab:appendix-rejudge-cases} presents two representative rejudging cases. The first case shows that the initial judge can be overly literal. For the checklist question \emph{``Does someone adjust the position of @Image2?''}, several outputs were initially scored as failures because no explicit ``adjusting'' action was identified. However, picking up, carrying, or placing the plant changes its position and satisfies the semantic intent of the checklist item. The rejudge stage therefore corrects these overly strict judgments. The second case illustrates the opposite failure mode. For \emph{``Is the scene set backstage of a theater?''}, some initial scores were too high because the judge accepted a general theater-stage setting as backstage. The rejudge stage lowers these scores when the evidence indicates a main stage or an ambiguous theater context rather than a clear backstage setting. Together, these cases show that rejudging improves consistency by aligning scores with both the intended checklist semantics and comparable visual evidence across models.

\begin{table}[t]
\centering
\small

\begin{tabular}{p{4.2cm}lccp{8.0cm}}
\toprule
\textbf{Question} & \textbf{Model} & \textbf{Before} & \textbf{After} & \textbf{Rejudge Rationale} \\
\midrule
\multirow{3}{3.6cm}{Does someone adjust the position of @Image2?}
& Kling 3.0& 1.0 & 5.0
& Picking up and placing the plant in a new location effectively changes its position, satisfying the intended action. \\
& Seedance 2.0 & 1.0 & 5.0
& The woman picks up, carries, and places the plant, which should be treated as a valid position adjustment. \\
& Gemini-Omni& 1.0 & 4.0
& The plant is picked up and held, which partially changes its position, although it is not clearly placed in a new location. \\
\midrule
\multirow{3}{3.6cm}{Is the scene set backstage of a theater?}
& Gemini-Omni & 1.0 & 2.0
& The video shows a theater main-stage context but lacks clear backstage evidence, such as off-stage props, exposed structures, or backstage space. \\
& Seedance 2.0 & 5.0 & 3.0
& The original reason describes the scene as ``on or backstage,'' indicating ambiguity rather than a clear backstage setting. \\
& Wan 2.7& 4.0 & 2.0
& The scene suggests a theater environment, but the evidence points to the main stage rather than a distinct backstage area. \\
\bottomrule
\end{tabular}
\caption{Representative rejudging cases. The rejudge stage corrects both overly strict and overly permissive initial judgments by checking whether the score is consistent with the visual evidence and the checklist semantics.}
\label{tab:appendix-rejudge-cases}
\end{table}

\subsection{Stability of MLLM-Dependent Evaluation}
To assess evaluation stability, we randomly sample 60 examples from the three main boards and repeat the complete MLLM-dependent evaluation pipeline five times using the same generated outputs and evaluation settings. EF and SLS are also reported because they incorporate MLLM-assisted processing: EF uses an MLLM-estimated paste-naturalness factor for post-hoc calibration, while SLS uses MLLM-based pre-screening to identify samples suitable for reliable lip-sync measurement. 

 As summarized in Table~\ref{tab:mllm-stability}, the relative standard deviations remain below 1\% across all metrics, while the item-level MAEs are consistently small. These results suggest that the reported evaluation outcomes are not artifacts of judge stochasticity and that the overall model comparisons remain reliable under repeated evaluation.

\begin{table}[t]
\centering
\small

\setlength{\tabcolsep}{4pt}
\renewcommand{\arraystretch}{1.12}
\begin{tabularx}{\columnwidth}{
@{}
>{\centering\arraybackslash}m{0.25\columnwidth}
>{\centering\arraybackslash}m{0.11\columnwidth}
>{\centering\arraybackslash}X
>{\centering\arraybackslash}m{0.15\columnwidth}
@{}}
\toprule
\textbf{Dimension} & \textbf{Metric} & \textbf{Score} & \textbf{MAE} $\downarrow$ \\
\midrule

\multirow{1}{*}{\shortstack[c]{Basic Quality}}
& AQ
& $3.655 \pm 0.022$
& 0.137 \\
\midrule

\multirow{3}{*}{\shortstack[c]{Reference\\Consistency}}
& EF
& $0.598 \pm 0.005$
& 0.014 \\
& BC
& $4.382 \pm 0.035$
& 0.094 \\
& DP
& $4.417 \pm 0.027$
& 0.101 \\
\midrule

\multirow{3}{*}{\shortstack[c]{Audio-Visual\\Consistency}}
& SLS
& $3.121 \pm 0.014$
& 0.082 \\
& ESM
& $4.768 \pm 0.031$
& 0.067 \\
& SC
& $4.846 \pm 0.009$
& 0.056 \\
\midrule

\multirow{4}{*}{\shortstack[c]{Instruction\\Following}}
& VT
& $4.274 \pm 0.039$
& 0.116 \\
& AT
& $4.626 \pm 0.046$
& 0.098 \\
& SCA
& $4.712 \pm 0.014$
& 0.087 \\
& TO
& $4.221 \pm 0.030$
& 0.128 \\
\bottomrule
\end{tabularx}
\caption{\textbf{Stability of MLLM-dependent evaluation over five repeated runs on a 60-sample subset.}
Score reports the mean and standard deviation of the five run-level average scores. MAE denotes the average item-level absolute difference over all pairs of runs.}
\label{tab:mllm-stability}
\end{table}

\section{Detailed Metric Constructions and Prompts}

This section details the construction of all evaluation metrics, including the generation of metric-specific checklists, the prompts used for MLLM-as-a-Judge evaluation, and the implementation details of automatic metrics.
For readability, the prompt boxes below present the core components of each evaluation prompt, while omitting output-format specifications and implementation-specific serialization details.

\subsection{Basic Quality}
\subsubsection{Anatomical Quality (AQ)}

\promptinput{Anatomical and Object Integrity Judge Prompt}{reference/all_prompts_pack/files_condensed/BQ/anatomical_knowledge.md}{1}{27}

\subsection{Reference Consistency}
\label{app:ef}
\subsubsection{Entity Fidelity (EF)}

Entity Fidelity (EF) measures whether the entities in a generated video remain visually faithful to their corresponding reference images. The metric combines automated reference-similarity branches with MLLM-based diagnostic checks. For each generated video, we first parse the sample metadata, load the corresponding human, object, and scene references, uniformly sample 32 frames, and compute branch-level similarity scores.

For human fidelity, we adopt different pipelines for real and cartoon subjects. For real-human references, the face branch detects faces using InsightFace and extracts ElasticFace R100 embeddings. The body branch detects person regions with YOLO-World and uses SigLIP embeddings as the final body similarity score. When both face and body evidence are available, the frame-level real-human score combines them with a default face weight of 0.6. For cartoon subjects, we use a YOLOv8 anime-face detector for the head branch and compute DINOv2-based head similarity, since real-face recognition models are unreliable for stylized faces. The cartoon body branch also uses YOLO-World crops and SigLIP embeddings as the final body score. Before combining head and body evidence, low-quality cartoon body matches with similarity below 0.6 are filtered out.
For object fidelity, we use YOLO-World open-vocabulary detection to localize referenced objects in sampled video frames and compute SigLIP similarity between the cropped detections and the reference object image. For scene fidelity, we compare full-frame SigLIP embeddings with the scene reference and use the best-matching sampled frame, since this branch targets global environment similarity rather than localized object appearance.

To account for temporal drift, the automated EF score is further adjusted by Long-Term Consistency (LTC). Given the frame-level EF sequence, LTC measures score fluctuation, collapse ratio, and late-stage degradation relative to early frames, and converts these factors into a stability coefficient in [0, 1]. The adjusted score is defined as
\[
\mathrm{EF}_{\mathrm{base}} =
\overline{\mathrm{EF}} \times (0.7 + 0.3 \times \mathrm{LTC})
\]
where \(\overline{\mathrm{EF}}\) denotes the mean automated EF score over sampled frames. Thus, videos with stable identity and appearance over time keep scores close to the original mean, while videos with identity drift, disappearance, or late-frame degradation receive a lower EF score.

In addition, the paste-naturalness coefficient is used as a correction to automated EF by assigning a visual-integration penalty to videos where the subject appears sticker-like. Overall, automated EF captures reference similarity, LTC captures temporal stability, and the MLLM components provide semantic and compositional diagnostics when high embedding similarity alone is insufficient.
The final score is:
\[
    \mathrm{EF}_{\mathrm{final}} = \mathrm{EF}_{\mathrm{base}} \times P_{\mathrm{paste}}(r)
\]

\promptinput{Compositing Artifact Judge Prompt}{reference/all_prompts_pack/files_condensed/EF/paste_artifact.md}{1}{39}

\subsubsection{Detail Preservation (DP)}

\promptinput{Fine-Grained Detail Preservation Judge Prompt}{reference/all_prompts_pack/files_condensed/EF/detail_preservation.md}{1}{48}

\subsubsection{Binding Correctness (BC)}

\promptinput{Binding Correctness Judge Prompt}{reference/all_prompts_pack/files_condensed/EF/binding.md}{1}{37}

\subsection{Audio-Visual Consistency}

\subsubsection{Speech-Lip Synchronization (SLS)}
For speech-lip synchronization, we adopt a two-stage evaluation pipeline. We first apply an MLLM-based pre-filter to retain only videos in which audible speech is produced by a visible speaker with stable frontal or near-frontal face visibility and sufficient mouth visibility. This filtering step reduces unreliable SyncNet measurements caused by profile faces, occlusions, or off-screen speech. For the retained videos, we apply a SyncNet-based VSLS pipeline on speech windows detected by voice activity detection (VAD). In multi-person scenes, candidate face tracks are first filtered according to their temporal offset. We then select the track with the lowest LSE-D as the speaking face, using lower absolute offset and higher LSE-C as tie-breakers. The selected track is further verified by a crop-level stable-frontal check to ensure that the synchronization score is computed on reliable facial evidence. For each valid speech window, we normalize LSE-C as a confidence score and reverse-normalize LSE-D as a distance score. The final speech-lip synchronization score is computed as \[ \mathrm{VSLS}_{1\text{-}5} = 1 + 4 \times \left(0.60 \cdot C_{\mathrm{norm}} + 0.40 \cdot D_{\mathrm{norm}}\right), \] where \(C_{\mathrm{norm}}\) denotes the normalized LSE-C score and \(D_{\mathrm{norm}}\) denotes the reverse-normalized LSE-D score. The resulting score is mapped to a 1--5 scale, where higher values indicate better temporal synchronization between speech and visible lip motion.

\promptinput{Stable Speaking Frontal Face Filtering Prompt}{reference/all_prompts_pack/files_condensed/AC/stable_speaking_frontal_face.md}{1}{37}

\subsubsection{Timbre Similarity (TS)}
\label{app:ts}
Timbre Similarity evaluates whether the generated speech preserves the speaker timbre specified by the voice reference. For each applicable sample, we first isolate the generated speech segment and the reference-voice segment, and then extract speaker embeddings from both clips using the same speaker-verification backbone. The raw sample-level similarity is computed as the cosine similarity between the generated and reference speaker embeddings, after filtering out clips with insufficient speech duration (>2 s), severe overlap with non-speech audio, or failed voice-activity detection.

To calibrate the scoring range, we synthesize auxiliary speech pairs with the same speaker timbre but different spoken content. We observe that their embedding cosine similarities typically fall in the range of 0.30 to 0.45, indicating that content variation can substantially lower the absolute similarity even when the perceived timbre remains consistent. Therefore, we treat a cosine similarity above 0.30 as sufficiently strong evidence of timbre preservation and assign it the maximum score. For similarities below this threshold, we linearly map the cosine similarity to a 1–5 scale, where higher scores indicate stronger timbre consistency. 

\subsubsection{Event-Sound Matching (ESM)}

\promptinput{Visual Event-Sound Alignment Judge Prompt}{reference/all_prompts_pack/files_condensed/AC/visual_event_sound_alignment.md}{1}{31}

\subsubsection{Source Correctness (SC)}

\promptinput{Source Correctness Judge Prompt}{reference/all_prompts_pack/files_condensed/AC/source_correctness.md}{1}{29}
\subsubsection{Timbre Similarity (TS).}
For voice timbre similarity, we use an embedding-based automatic evaluation protocol. For each sample with a voice reference, we extract the generated audio, convert both reference and generated speech to 16 kHz mono waveforms, and apply energy-based VAD to retain valid speech regions. Samples with less than 2 seconds of speech are excluded. We then use the SpeechBrain ECAPA-TDNN speaker embedding model to extract speaker embeddings and compute cosine similarity between the reference and generated speech. Since generated videos often contain short utterances, background audio, sound effects, and content different from the reference, the raw cosine values are lower than those in clean speaker verification. We therefore calibrate the score mapping using synthesized speech clips with the same timbre but different content, where cosine similarities typically fall between 0.30 and 0.45. Based on this observation, we treat cosine similarity above 0.30 as strong timbre consistency and map cosine scores to a 1--5 scale with thresholds of 0.075, 0.150, 0.225, and 0.300. The final model-level score is averaged over all valid samples.

\subsection{Instruction Following}

\paragraph{Checklist generation prompt.}
\promptinput{Audio, Speech, and Temporal Checklist Generation Prompt}{reference/all_prompts_pack/files_condensed/IF/if_checklist_generation.md}{1}{32}

\subsubsection{Visual Task Following (VT)}

\promptinput{Visual Checklist Generation Prompt}{reference/all_prompts_pack/files_condensed/IF/visual_checklist_generation.md}{1}{27}

\promptinput{Visual Task Following Judge Prompt}{reference/all_prompts_pack/files_condensed/IF/visual_task_following.md}{1}{33}

\subsubsection{Audio Task Following (AT)}

\promptinput{Audio Task Following Judge Prompt}{reference/all_prompts_pack/files_condensed/IF/audio_task_following.md}{1}{26}

\subsubsection{Speech Content Accuracy (SCA)}

\promptinput{Speech Content Accuracy Judge Prompt}{reference/all_prompts_pack/files_condensed/IF/speech_accuracy.md}{1}{33}

\subsubsection{Temporal Order Following (TO)}

\promptinput{Temporal Order Following Judge Prompt}{reference/all_prompts_pack/files_condensed/IF/temporal_order_following.md}{1}{36}
